\documentclass{article}
\usepackage[utf8]{inputenc}
\usepackage[T1]{fontenc}

\usepackage{arxiv}

\usepackage[numbers]{natbib}

\usepackage{geometry}
\usepackage{graphicx}
\graphicspath{{./Images/}}
\usepackage{array}
\usepackage{multirow}
\usepackage{booktabs}
\usepackage{float}

\usepackage{amsmath, amsfonts}
\usepackage{nicefrac}
\usepackage{microtype}
\usepackage{enumitem}

\usepackage{pgfplots}
\pgfplotsset{compat=1.18}

\usepackage[ruled,vlined]{algorithm2e}

\usepackage{url}
\usepackage[unicode]{hyperref}
\hypersetup{
  colorlinks=true,
  linkcolor=blue,
  citecolor=blue,
  urlcolor=blue
}

\title{ArbESC+: Arabic Enhanced Edit Selection System Combination for Grammatical Error Correction \\ Resolving conflict and improving system combination in Arabic GEC}

\author{
  Ahlam Alrehili \\
  Department of Computer Sciences,\\ Faculty of Computing and Information Technology  \\
  King Abdulaziz University, Saudi Electronic University \\
  Medina, Saudi Arabia \\
  \texttt{a.alrehili@seu.edu.sa} 
  \And
  Areej Alhothali \\
  Department of Computer Sciences,\\ Faculty of Computing and Information Technology \\
  King Abdulaziz University \\
  Jeddah, Saudi Arabia \\
  \texttt{aalhothali@kau.edu.sa}
}

\begin{document}
\maketitle
\begin{abstract}
Grammatical Error Correction (GEC) is an important aspect of natural language processing. Arabic has a complicated morphological and syntactic structure, posing a greater challenge than other languages. Even though modern neural models have improved greatly in recent years, the majority of previous attempts used individual models without taking into account the potential benefits of combining different systems. In this paper, we present one of the first multi-system approaches for correcting grammatical errors in Arabic, the Arab Enhanced Edit Selection System Complication (ArbESC+). Several models are used to collect correction proposals, which are represented as numerical features in the framework. A classifier determines and implements the appropriate corrections based on these features. In order to improve output quality, the framework uses support techniques to filter overlapping corrections and estimate decision reliability. A combination of AraT5, ByT5, mT5, AraBART, AraBART+Morph+GEC, and Text editing systems gave better results than a single model alone, with F0.5 at 82.63\% on QALB-14 test data, 84.64\% on QALB-15 L1 data, and 65.55\% on QALB-15 L2 data. As one of the most significant contributions of this work, it's the first Arab attempt to integrate linguistic error correction. By improving existing models it provides a practical step towards developing advanced tools that will benefit users and researchers of Arabic text processing.
\end{abstract}

% keywords can be removed
\keywords{Grammatical error correction \and GEC \and System Combination \and QALB \and ESC}

\section{Introduction}
Correcting grammatical errors improves both the clarity and precision of the text by identifying and eliminating linguistic, morphological, or spelling errors~\cite{ng2025conll}. GEC systems have made significant progress for resource-rich languages, such as English; however, the development of GEC systems for Arabic remains challenging. A combination of highly inflected morphology, diverse orthographic conventions, and complex syntactic structures of the language, as well as the scarcity of large and high-quality annotated corpora, makes it extremely difficult for machine learning models to perform well. Recent advances in synthetic data generation techniques have improved Arabic grammar error correction systems, such as~\cite{solyman2021synthetic},~\cite{solyman2023optimizing},~\cite{alrehili2025towards}, and~\cite{alrehili2025tibyan}. The synthetic data generation techniques enabled us to develop rich training environments that simulate realistic error patterns while embracing a broad spectrum of grammatical forms and structures. This improved accuracy and applicability despite the lack of Arabic-language resources for the models.

A transformer architecture is demonstrated to be efficient in natural language processing, whether it is based on sequence-to-sequences~\cite{vaswani2017attention} or sequence labeling~\cite{stahlberg2020seq2edits}. A modern system relies heavily on its capability to represent linguistic context and comprehend complex relationships between words and structures. It has become common to develop prompts for text to generate grammatical corrections using large language models (LLMs), such as GPT-4. Therefore, a wide range of new research approaches has emerged for improving output quality, accuracy, and expanding the development of advanced language systems.

A major improvement to GEC systems has been made by text-to-text transformers such as AraT5~\cite{nagoudi2021arat5} and its multilingual variants. Recent research has proven that AraT5 is a powerful model in the field of Arabic GEC~\cite {kwon2023chatgpt}~\cite{alrehili2025towards}~\cite{ismail2025transformers}. Although multilingual models such as BYT5~\cite{xue2022byt5} and mT5~\cite{xue2020mt5} have shown significant success in many low-resource languages~\cite{rothe2021simple}, they remain underexplored in the Arabic context, despite being effective in processing raw text and handling orthographic and morphological variations in many low-resource languages~\cite{ingolfsdottir2023byte}.

The quality of the GEC system can be improved by combining the output of single-model systems. It has become increasingly common to use model fusion techniques to improve GEC systems' performance. The most common approaches are ensembling and system combination~\cite{bryant2023grammatical}. Typically, ensembling combines the outputs of models, such as voting or weighted averaging, whereas system combination allows for a deeper level of integration, combining editing levels from different systems, allowing for more accurate selection of corrections~\cite{kantor2019learning}. Despite the success of the system combination approach in correcting grammatical errors in multiple languages, this approach has not been documented in Arabic, which opens up an important and unexplored field for research.

To address the gap, we propose the ArbESC+ framework, which combines outputs from models trained specifically for Arabic. ArbESC+ will contribute to taking advantage of the diversity of the models, incorporating Arabic morphology and syntax into a more accurate and comprehensive correction of grammatical errors. This framework is used in two integrated stages to maximize the benefits of the diversity of models and data sources. The first stage involves fine-tuning three text-to-text transformer-based models, AraT5, AraBart, ByT5, and mT5, to fit the unique characteristics of grammatical and spelling errors in Arabic texts. Additionally, we extract the output from the AraBART+Morph+GEC~\cite{alhafni2023advancements} model, which contains three versions that have been trained on QALB-14, QALB-15, and zeabuq, and text-editing\cite{alhafni2025enhancing} has two versions, one that has been trained on QALB-14 and one that has been trained on zeabuq. As a result, we have 9 models. Text editing and AraBART+Morph+GEC models are used off-the-shelf without retraining, leveraging their existing capabilities to process Arabic texts. During the second stage, an enhanced edit selection combination mechanism is used to combine all outputs at the micro-edit level, using conflict resolution strategies specifically tailored to Arabic grammatical structures and spelling patterns, ensuring the most accurate and contextually consistent corrections.

Our contributions are as follows.
\begin{itemize}
    \item Comparison of GEC methods in a comprehensive manner. Reproduction, evaluation, and comparison of the most promising existing GEC methods are presented, including single-model systems as well as ensembles. In this paper, we show that the combination of systems plays a crucial role in obtaining state-of-the-art performance in GEC.

    \item The establishment of new state-of-the-art baselines. In our study, we demonstrate that a simple, easy system combination outperforms more complex approaches and boosts performance significantly. We are pushing the boundaries of GEC quality and achieving state-of-the-art results on QALB-14 and QALB-15.

    \item A novel approach to combining grammatical error correction systems is proposed by formulating the task as a binary classification that predicts each edit independently. To our knowledge, this is the first system combination for grammatical error correction in Arabic.

    \item A revised Easy system combination (ArbESC+), including advanced conflict resolution strategies, is developed to address Arabic grammatical structures and orthographic patterns.

    \item  ArbESC+ can be generalized to low-resource languages, enabling its application in multiple language environments that face similar challenges in terms of data scarcity and complex linguistic structures.
\end{itemize}

\section{Related Work}
The following sections provide an overview of the literature related to the Arabic GEC. We will discuss models and techniques developed to address the challenges associated with correcting grammatical errors in Arabic. Moreover, we will discuss a study that examined the use of system combinations to improve the accuracy and reliability of English texts.

\subsection{Arabic Grammatical Error Correction Model}

There has been an increase in interest in exploiting T5 models for a variety of natural language processing tasks in the Arabic language in recent years. A pre-trained model developed by Nagoudi et al.~\cite{nagoudi2021arat5} with extensive Arabic corpora outperforms multilingual models in the Arabic generation (ARGEN) and Arabic language understanding evaluation (ARLUE) benchmarks in tasks such as generation and comprehension. Further research has demonstrated AraT5's superior performance on Arabic grammatical error correction, particularly on the QALB-2014 benchmark. Meanwhile, Kwon et al.~\cite{kwon2023chatgpt} investigated how to correct Arabic grammatical errors using ChatGPT and GPT-4. In their study, it was found that GPT-4 achieved competitive results F0.5=67\%, but its performance fell behind specialized Arabic models such as AraT5 and AraBART. Comparatively, AraT5 reported F0.5 scores exceeding 75\% on the QALB-2014 dataset, demonstrating the benefit of domain-specific pretraining and task-oriented fine-tuning. However, Arabic-specific models still outperform large models regarding corrective performance. On the QALB-2014, Alhafni et al.~\cite{alhafni2023advancements}  found that AraBART and AraT5 achieved competitive performance. They fine-tuned the AraBART and AraT5 models for each QALB-2014, QALB-2015, and ZAEBUC datasets. Three models were developed: AraBART and AraT5, AraBART+Morph and AraT5+Morph, and AraBART+Morph+GED and AraT5+Morph+GED, which combined morphological parsing with syntactic error detection. Based on the results, AraBART+Morph+GED outperformed the baseline model significantly. The AraBART+Morph+GED model scored F0.5 with 79.6\% in the QALB-2014 test set and 80.3\% in QALB-15, resulting in the model of the best performer. On the other hand, Alrehili and Alhothali~\cite{alrehili2025towards} show that AraT5 is capable of correcting grammatical errors based on syntactic data by scoring F0.5 79.36\% in the QALB-14. Abdelrehimhybrid et al.~\cite{abdelrehimhybrid} show the performance of AraBART fine-tuned on their synthetic data, achieving F0.5 is 80.1\% on the QALB-2015 test set with punctuation and 88.6\% without punctuation. 

Models based on Seq2Seq have shown an ability to correct errors by generating fully corrected sentences. Therefore, this approach can be time-consuming and costly, and it can sometimes result in unnecessary text changes. Sequence Labeling models are able to identify errors more accurately and suggest only modifications that are necessary, thus increasing their efficiency. There has been extensive use of sequence labeling in English studies, but this technique is not widely used in Arabic studies. Sequence labeling is introduced to GEC tasks for the first time in Alhafni et al.~\cite{alhafni2025enhancing}. A text editing approach was proposed as an alternative method to correct grammatical errors in Arabic rather than sequence-to-sequence models. A sentence is viewed as a series of linguistic units that can be rearranged simply by retaining, deleting, replacing, or inserting. Researchers compressed edit patterns, used subword representations, and separated punctuation errors from other errors to deal with Arabic's complex morphological structure. It was demonstrated that this methodology achieves faster, more accurate results than the best state-of-the-art models for MADAR CODA, QALB, and ZAEBUC data set. According to the QALB-14 and 15 tests, they achieved an F0.5 score of 80.3\% and 80.5\%, respectively.

In addition, Alhafni and Habash contribute ~\cite{alhafni2025enhancing} a voting mechanism to the process of combining the outcomes of several scoring systems. Overall, ensemble models performed better than individual models across all datasets, showing that this strategy improved performance. The ensemble system achieved F0.5 equal to 81.7\% on the QALB-2014 corpus, 82.9\% on the QALB-2015 corpus, and 87.2\% on the ZAEBUC corpus. Ensemble system performed best on dialect data (MADAR CODA), 88.9\%.

Arabic language development is based on AraT5 and its variations, according to research studies. The accuracy of Arabic grammar has improved significantly, but there are still limitations in the field. There has been little exploration of the ByT5 model in the Arabic GEC, despite its success in addressing grammatical errors in languages with some morphological similarities to Arabic~\cite{ingolfsdottir2023byte}. As a result, sequence labeling methodologies have been relatively under-utilized in Arabic, despite their efficiency and speed. They are limited in their ability to correct deep and complex errors, despite their proven effectiveness in English. While model fusion methods have demonstrated a clear ability to improve accuracy~\cite{alhafni2025enhancing}, they are still challenged by high computational costs and the difficulty of achieving complete consistency.

\subsection{System Combination}
In this section, we review some of the prior work on combining GEC systems. Jayaraman et al.~\cite{jayaraman2005multi} developed the Multi-Engine Machine Translation Guided by Explicit Word (MEMT) method to incorporate machine translation hypotheses from different base systems. The MEMT approach also works for GEC system combinations acording to Susanto et al.~\cite{susanto2014system}. Using MEMT, the hypotheses are first aligned, and then all candidate token paths are generated from the hypotheses. As MEMT searches for potential tokens, specific constraints apply, such as monotonicity, completion, and no repetition. The MEMT scores the proposed tokens based on n-gram scores, the token number, and the hypothesized n-gram similarity.

The Iterative Best Match (IBM) technique developed by Kantor et al.~\cite{kantor2019learning} categorizes edits from two hypotheses into three datasets: edits present only in one hypothesis, edits present in only the second hypothesis, and edits contained in both hypotheses. For each edit type, these datasets are created. Next, the model determines which dataset to include for every edit type. It only combines two systems at a time. Therefore, when combining more than two systems, this method must be applied iteratively.

Based on Lin and Ng 's~\cite{lin2021system}, the Grammatical Error Correction–Integer Programming (GEC-IP) system is similar to IBM, but it directly optimizes and simplifies parameters instead of optimizing real-valued parameters and rounding them later, as with IBM. Moreover, GEC-IP can combine more than two base systems simultaneously, as opposed to combining only two at a time with IBM. According to GEC-IP, for each edit type, only one base system's edits are used as final corrections, ignoring edits from the other systems.

Diversity-driven combinations (DDC)~\cite{lin2021system} aim to produce a more effective combined system by increasing diversity among the base systems. A useful combination of base systems must be diverse (almost uncorrelated) and of similar quality, according to Macherey and Och~\cite{macherey2007empirical}. A DDC system cannot be completely black-boxed, as it requires a basis system that can be fine-tuned to function as the backbone. In DDC, reinforcement learning creates diversity in the base systems before using an off-the-shelf system combination method.

Edit-based System Combination (ESC)~\cite{qorib2022frustratingly} combines hypotheses from different grammatical error correction base systems by formulating each edit as an independent binary classification. A logistic regression model was used to classify the data. They selected the edits based on probabilities obtained from the model using a greedy strategy. Initially, they only considered edits with probabilities above a certain threshold. After that, the edits were sorted from highest to lowest probability and checked one by one to ensure they did not conflict with previously selected edits.

Among the three approaches (IBM, GEC-IP, MEMT), one used only output sentences, and the other two used output sentences and proposed edit types based on hypotheses (output sentences of component systems). Additionally, in this section, we discuss the ESC, which introduces diversity to the base systems, which complement our method of combining base systems.

In conclusion, most of the research on the Arabic language has focused on developing individual model-based corrections for grammatical errors, while studies on system combinations have remained scarce. The ESC approach has made a significant contribution to the English language in this regard, but it remains limited in terms of how it handles conflicts between outputs. Most models are compared by simple voting or weighting instead of taking into account the differences between them in depth. As a result, our contribution offers a direct improvement over this approach by introducing a sophisticated mechanism for dealing with conflicts between outputs, thus maximizing the strengths of each model while reducing the negative effects of differences. Hence, our work represents the first systematic attempt to employ system combinations for Arabic grammatical correction, with conflict resolution as one of the most important challenges.
\section{Single-Model Systems}
This section discusses several grammatical and linguistic error correction models available for Arabic, which are divided into two categories: Seq2Seq models and Edit-based models.
\subsection{Sequence-to-Sequence models}
In terms of GEC, Seq2Seq models are among the most popular and effective techniques. Seq2seq models can transform an input sentence, which may contain errors, into a corrected one. This approach provides accurate corrections for spelling, grammar, and morphology errors, as well as improving the stylistic coherence of a text by considering the full context of the sentence. In the following sections, four of the most popular T5~\cite{raffel2020exploring} models will be discussed: AraT5~\cite{nagoudi2021arat5}, ByT5~\cite{xue2022byt5}, and mT5~\cite{xue2020mt5}, and their ability to handle and correct linguistic errors in Arabic, including grammatical, morphological, and spelling errors. 

\subsubsection{AraT5}
The AraT5\cite{nagoudi2021arat5} is an Arabic language model based on the T5~\cite{raffel2020exploring} architecture, pre-trained on large Arabic text datasets.  The model version (AraT5v2-base-1024) was used from the Hugging Face repository and fine-tuned using QALB-2014, QALB-2015, and zeabuq datasets jointly to improve its performance in correcting syntactic, morphological, and spelling errors in Arabic texts. In this study, training was carried out for 150 epochs with 8 batch sizes and 2e-5 initial learning rates, with mixed precision (fp16) enabled, AdamW optimization, and default learning schedule settings. In order to enhance the model's ability to distinguish between correct and incorrect texts, the inputs were prefixed with "grammar:". This prefix was used to guide the model during training.
\subsubsection{ByT5}
The ByT5~\cite{xue2022byt5} is an implementation of the T5~\cite{raffel2020exploring} framework with a sequence-to-sequence architecture. It processes text directly at the byte level, which makes it more adept at handling spelling and typographical variations more effectively. We used Hugging Face's byt5-base version as a starting point and fine-tuned it by prefixing model inputs with "grammar:" to match the QALB-2014 and QALB-2015 datasets.

The input and reference texts were prepared so that the length of their text sequences did not exceed 256, and the maximum generation length was set to the same value during evaluation. We trained over 150 epochs, with early stopping enabled by periodic evaluations and loading the best model at the end. A total of eight batches were produced per machine, with four gradient accumulation steps used to increase the effective batch size. We used a ByT5Tokenizer that is appropriate for processing literal texts, along with a dataset that is compatible with the sequencing task.

\subsubsection{mT5}
mT5~\cite{xue2020mt5} is a multilingual generative model, which was initially trained using datasets covering more than 100 languages. It was not specifically optimized for Arabic. While its generalization capabilities can be fine-tuned, it maintains good generalization capabilities. We fine-tuned the MT5-base starting point from Hugging Face with the QALB-2014 and QALB-2015 datasets without adopting a significant prefix (the phrase "grammar:" is not added to the input file). We prepared the data using uniform input and reference encoding with truncation/padding up to 512 characters. A suitable sequence dataset was used for generation with alignment to multiples of 8.

During training, 150 epochs were run at a learning rate of 2e-5, with 4 gradient accumulation steps per machine and 4 batch sizes. A periodic evaluation strategy was implemented every 500 steps, with checkpoints saved every 1000 steps and logs tracked every 10 steps. There was no explicit option to enable label normalization or early stopping. 
\subsubsection{AraBART}
The BART model~\cite{lewis2019bart}, incorporating an encoder–decoder architecture that can reconstruct input texts that have been hidden or distorted. This model inspired the development of the Arabic version of BART \cite{antoun2020arabert}, which represents the first adaptation of BART specifically for the Arabic language. AraBART \cite{antoun2020arabert} was trained on large Arabic corpora to capture its morphological and syntactic characteristics. It has proven effective in translating, paraphrasing, and correcting grammatical errors. AraBART has demonstrated competitive performance in Arabic Grammar Error Correction (Arabic GEC) when trained on real-world datasets such as QALB-2014 \cite{mohit2014first} and QALB-2015 \cite{rozovskaya2015second}. According to its comparisons, it excels at identifying complex errors requiring a thorough understanding of the text's syntactic context, while simple errors of a spelling or morphological nature are processed in a manner consistent with other models such as AraT5 \cite{nagoudi2021arat5} and mT5 \cite{xue2020mt5}.

AraBART was tuned on QALB data with pairs of originals and corrections. Training efficiency was improved by encoding texts to 512 tokens with padding optimization. We used a low learning rate (2e-5) and a batch size of approximately 16, and we trained 50 epochs.
\subsubsection
  [AraBART+Morph+GEC]%
  {AraBART+Morph+GEC \texorpdfstring{\protect\footnotemark}{}}%
\footnotetext{\url{https://github.com/CAMeL-Lab/arabic-gec}}
The AraBART+Morph+GEC13 \cite{alhafni2023advancements} grammatical error correction model was developed by fine-tuning the AraBART model using QALB-2014, QALB-15, and Zeabuq data. Morphologically pre-processed texts were used to train the model hence that it could handle morphological variations in Arabic better.

There are three versions of the AraBART+Morph+GEC model available on the Hugging Face platform: the AraBART+Morph+GEC13 QALB-2014 Model, the AraBART+Morph+GEC13 QALB-2015 Model, and the AraBART+Morph+GEC13 ZAEBUC Model. The models were run using the same settings as the reference code included in the official repository, which relies on Transformers from Hugging Face as well as morphological processing tools from CAMeL Tools.  Grammatical Error Detection (GED) was used to label the processed text to identify grammatical errors. Then, an optimized GEC model used these labels in conjunction with the input text to handle this additional information. In order to generate corrected texts, we used a beam search algorithm using five beams and a maximum generation length of 100, while maintaining the rest of the parameters. 
\subsection{Edit-based Systems}
Unlike traditional sequential models, the Edit-Based System (EBS) focuses on text processing by identifying, classifying, and correcting errors at the level of small units, such as words, syllables, or symbols, rather than regenerating entire sentences. An appropriate correction for each error location in the text is first proposed on the basis of the error type or grammatical category.
\subsubsection
  [Text Editing]%
  {Text Editing \texorpdfstring{\protect\footnotemark}{}}%
\footnotetext{\url{https://github.com/CAMeL-Lab/text-editing}}

In text editing~\cite{alhafni-habash-2025-enhancing}, edit tags are applied to input tokens when text editing applies GEC, resulting in corrected text. In this study, four grammatical error correction models, each of which is worked on in a complementary fashion, were used. The first two models correct non-punctuation errors (NoPnx) as well as punctuation errors (Pnx) in the QALB-2014 texts. These models are based on the AraBERTv02 model, which is trained with the SWEET-Subword Edit Error Tagger approach. The second pair is based on the same idea, but it uses ZAEBUC data instead of QALB. First, all grammar and spelling errors except punctuation are corrected using the NoPnx model. The resulting text is then passed to the Pnx model for punctuation modification. The sequence allows specialized processing of each type of error, resulting in a higher degree of accuracy compared to using a single, comprehensive model.

\subsection{Single-Model Systems Results}
According to Table \ref{tab:qalb-results}, models perform differently depending on the dataset. According to the QALB-14-L1 Test set, the Text-edit-qalb14 model achieved the highest precision 84.14\%, but the lowest recall 67.0\%. There was a balance between precision and recall, resulting in a high F0.5 80.05\% but not significantly outperforming AraT5 80.09\%, with a better balance between precision and recall 83.10\%. The BYT5 model had a lower F0.5 75.94\% than the others, with a decline in all metrics.

AraT5 achieved the highest F0.5 at 80.06\% in the QALB-L1-15 test, while Text-edit-qalb14 followed with 80.04\%. Comparatively, AraBART+Morph+GEC13 (2015) and AraBART+Morph+GEC (zeabuc) performed relatively well 77–78\% but not as well as modern Transformer models. QALB-L2-15 Test performance declined across all models, with F 0.5 values ranging from 45\% to 64\% on the most challenging assessment for learners' writing. The AraT5 model performed the best, achieving 64.22\%, clearly outperforming its closest competitor, AraBART+Morph+GEC13 (2015) 60.98\%. However, editing systems such as Text-edit-zeabuc showed a noticeable decline 45.99\% due to a sharp decline in recall 23.86\%.

The Table \ref{tab:qalb-dev-results} confirms previous findings. AraT5 scored the highest F1 76.29\% and F0.5 79.69\% on QALB-L1-14 DEV, outperforming MT5 77.67\% and AraBART 77.01\%. A balanced score lower than AraT5 was achieved by Text-edit-14, which achieved the highest precision 83.51\% and recall 68.64\%. QALB-L2-15 DEV, where the task difficulty increases, AraT5 continues to outperform F0.5 65.26\%, more than five points ahead of MT5 60.91\%. Alternatively, BYT5 and Text-edit-zeabuc showed severe weaknesses, with F0.5 not exceeding 56\% and 44\%, respectively. Although editing-based systems consistently demonstrated high accuracy, they suffered from a sharp decline in recall, which explains their limited performance overall.
\begin{table}[ht]
\centering
\renewcommand{\arraystretch}{1.15}
\setlength{\tabcolsep}{3.5pt}
\resizebox{\textwidth}{!}{%
\begin{tabular}{lcccccccccccc}
\toprule
\multirow{2}{*}{\textbf{Model}} & \multicolumn{4}{c}{\textbf{QALB-14-L1 Test}} & \multicolumn{4}{c}{\textbf{QALB-L1-15 Test}} & \multicolumn{4}{c}{\textbf{QALB-L2-15 Test}} \\
\cmidrule(lr){2-5} \cmidrule(lr){6-9} \cmidrule(lr){10-13}
 & P & R & F1 & F0.5 & P & R & F1 & F0.5 & P & R & F1 & F0.5 \\
\midrule
AraBART & 81.78\% & 66.44\% & 73.31\% & 78.17\% & 79.65\% & 71.89\% & 75.57\% & 77.97\% & 67.12\% & 43.30\% & 52.64\% & 60.46\% \\
AraT5 & 83.10\% & 69.97\% & 75.97\% & \textbf{80.09\%} & 81.14\% & 76.04\% & \textbf{78.50\%} & \textbf{80.06\%} & \textbf{71.29\%} & 45.98\% & \textbf{55.90\%} & \textbf{64.22\%} \\
BYT5 & 78.23\% & 68.00\% & 72.76\% & 75.94\% & 76.20\% & 72.69\% & 74.40\% & 75.47\% & 59.29\% & 40.29\% & 47.98\% & 54.18\% \\
MT5 & 82.84\% & 65.72\% & 73.29\% & 78.73\% & 81.00\% & 70.67\% & 75.49\% & 78.70\% & 67.46\% & 39.24\% & 49.61\% & 58.98\% \\
AraBART+Morph+GEC13 (2014)~\cite{alhafni2023advancements}  & 79.48\% & 65.41\% & 71.76\% & 76.20\% & 80.84\% & 71.76\% & 76.03\% & 78.85\% & 63.95\% & 29.76\% & 40.62\% & 52.00\% \\
AraBART+Morph+GEC13 (2015)~\cite{alhafni2023advancements} & 80.95\% & 68.02\% & 73.92\% & 77.99\% & 79.08\% & 73.80\% & 76.35\% & 77.97\% & 66.12\% & \textbf{46.52\%} & 54.61\% & 60.98\% \\
AraBART+Morph+GEC (zeabuc)~\cite{alhafni2023advancements} & 81.02\% & 67.31\% & 73.53\% & 77.85\% & 79.02\% & 72.98\% & 75.88\% & 77.73\% & 66.20\% & 45.81\% & 54.15\% & 60.79\% \\
Text-edit-qalb14~\cite{alhafni-habash-2025-enhancing} & \textbf{84.14\%} & 67.00\% & 74.60\% & 80.05\% & \textbf{81.91\%} & 73.33\% & 77.38\% & 80.04\% & 68.38\% & 27.95\% & 39.68\% & 53.04\% \\
Text-edit-zeabuc~\cite{alhafni-habash-2025-enhancing} & 81.91\% & \textbf{71.39\%} & \textbf{76.29\%} & 79.56\% & 78.50\% & \textbf{76.22\%} & 77.35\% & 78.04\% & 59.87\% & 23.86\% & 34.12\% & 45.99\% \\
\bottomrule
\end{tabular}%
}
\caption{Results single model systems evaluated on QALB-14, QALB-15-L1 and QALB-15-L2 test sets.}
\label{tab:qalb-results}
\end{table}

\begin{table}[ht]
\centering
\renewcommand{\arraystretch}{1.15}
\setlength{\tabcolsep}{3.5pt}
\resizebox{\textwidth}{!}{%
\begin{tabular}{lcccccccc}
\toprule
\multirow{2}{*}{\textbf{Model}} & \multicolumn{4}{c}{\textbf{QALB-L1-14 DEV}} & \multicolumn{4}{c}{\textbf{QALB-L2-15 DEV}} \\
\cmidrule(lr){2-5} \cmidrule(lr){6-9}
 & P & R & F1 & F0.5 & P & R & F1 & F0.5 \\
\midrule
AraBART & 80.16\% & 66.56\% & 72.73\% & 77.01\% & 66.03\% & 42.99\% & 52.07\% & 59.63\% \\
AraT5 & 82.13\% & 71.23\% & \textbf{76.29\%} & 79.69\% & \textbf{72.24\%} & \textbf{47.07\%} & \textbf{75.89\%} & \textbf{65.26\%} \\
BYT5 & 77.73\% & 68.98\% & 73.09\% & 75.80\% & 61.15\% & 41.99\% & 49.79\% & 56.04\% \\
MT5 & 81.13\% & 66.36\% & 73.00\% & 77.67\% & 68.86\% & 41.68\% & 51.93\% & 60.91\% \\
AraBART+Morph+GEC13 (2014)~\cite{alhafni2023advancements} & 81.09\% & 66.02\% & 72.79\% & 77.55\% & 63.30\% & 29.11\% & 39.88\% & 51.26\% \\
AraBART+Morph+GEC13 (2015)~\cite{alhafni2023advancements} & 80.17\% & 68.35\% & 73.79\% & 77.49\% & 64.78\% & 46.15\% & 53.90\% & 59.94\% \\
AraBART+Morph+GEC (ZEABUQ)~\cite{alhafni2023advancements} & 79.96\% & 67.47\% & 73.18\% & 77.10\% & 64.66\% & 43.93\% & 52.32\% & 59.09\% \\
Text-edit-14~\cite{alhafni-habash-2025-enhancing} & \textbf{83.51\%} & 68.64\% & 75.35\% & \textbf{80.04\%} & 66.50\% & 26.59\% & 37.99\% & 51.14\% \\
Text-edit-Zeabuq~\cite{alhafni-habash-2025-enhancing} & 80.47\% & \textbf{72.14\%} & 76.07\% & 78.65\% & 57.58\% & 22.41\% & 32.26\% & 43.82\% \\
\bottomrule
\end{tabular}%
}
\caption{Results single model systems evaluated on QALB-14, QALB-15-L2 development sets.}
\label{tab:qalb-dev-results}
\end{table}

\begin{figure}[ht]
\centering
\begin{tikzpicture}
\begin{axis}[
    ybar,
    bar width=12pt,
    width=\textwidth,
    height=8cm,
    ymin=65, ymax=80.5,
    ylabel={Macro Avg $F_{0.5}$ (\%)},
    symbolic x coords={
      AraT5,
      AraBART+Morph+GEC13 (2015),
      AraBART,
      MT5,
      AraBART+Morph+GEC (zeabuc),
      Text-edit-qalb14,
      AraBART+Morph+GEC13 (2014),
      BYT5,
      Text-edit-zeabuc
    },
    xtick=data,
    x tick label style={
      rotate=45,
      anchor=east,
      font=\small
    },
    nodes near coords,
    nodes near coords align={vertical},
    every node near coord/.append style={
      /pgf/number format/fixed,
      /pgf/number format/precision=2,
      font=\scriptsize
    },
    ymajorgrids,
    grid style={dotted},
    ylabel style={font=\bfseries},
    enlarge x limits=0.05,
    legend style={
      font=\small,
      at={(rel axis cs:0.02,0.98)}, % أعلى يسار داخل المحور
      anchor=north west,
      draw=none, fill=none
    }
]

% الأعمدة (من دون إدخالها في اللجند)
\addplot+[ybar, fill=gray!40, draw=gray!70, forget plot]
coordinates {
  (AraT5,74.79)
  (AraBART+Morph+GEC13 (2015),72.31)
  (AraBART,72.20)
  (MT5,72.14)
  (AraBART+Morph+GEC (zeabuc),72.12)
  (Text-edit-qalb14,71.04)
  (AraBART+Morph+GEC13 (2014),69.02)
  (BYT5,68.53)
  (Text-edit-zeabuc,67.86)
};

% خط المتوسط (مع إدخاله في اللجند فقط)
% خط المتوسط (مع ظهوره ودخوله في اللجند)
\addplot+[thick, dashed] coordinates {
  (AraT5,71.11)
  (Text-edit-zeabuc,71.11)
};
\addlegendentry{Mean}

\addlegendentry{Overall mean ($71.11\%$)}

\end{axis}
\end{tikzpicture}
\caption{Macro average $F_{0.5}$ across all QALB test sets. Labels are rotated to avoid overlap; the overall mean is shown as a dashed line with a legend.}
\label{fig:macro-f05}
\end{figure}
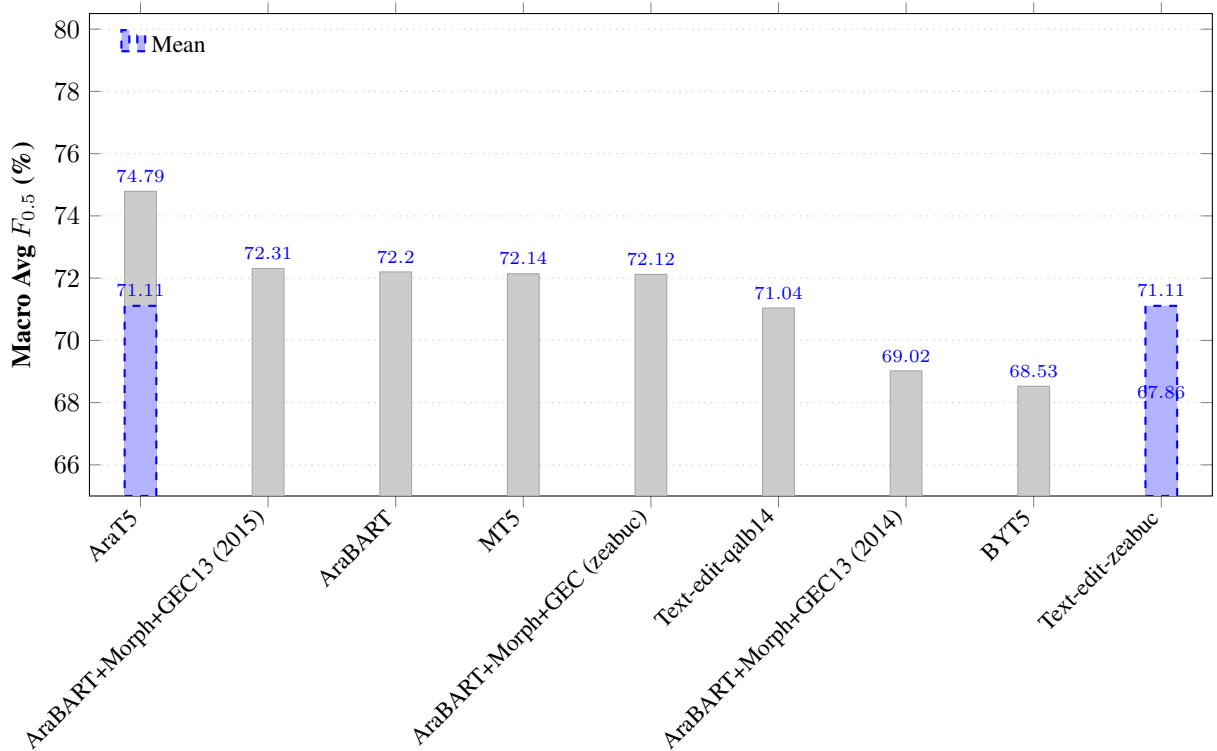

In terms of the overall average F0.5 across all test sets, Figure (\ref{fig:macro-f05}) shows that AraT5 is the most discriminatory, with an average of 74.79\%, nearly three points ahead of its nearest competitor (AraBART+Morph+GEC13 (2015) with 72.31\%). BYT5 and Text-edit performed below 70\%, while AraBART, MT5, and AraBART+Morph+GEC (zeabuc) achieved similar results (around 72\%). AraT5 is not only the best on certain sets, but also the most consistent across all evaluations, maintaining a good balance between precision and recall even in the most challenging L2 settings. Despite local strengths (such as high accuracy in Text-edit and flexibility in AraBART), other models did not achieve sufficient consistency across datasets.
% في الـ preamble
% \usepackage{pgfplots}
% \pgfplotsset{compat=1.18}

\section{Training and Evaluation Data}
\textbf{QALB-14 and QALB-15 }. Our training and evaluation sets were QALB-14~\cite{mohit2014first} and QALB-15~\cite{zhu2015aligning}. QALB-14 is composed of Al Jazeera News comments, written by native Arabic speakers. In QALB-2015, Arabic speakers of all backgrounds were addressed, including native Arabic speakers and nonnative Arabic speakers. This includes texts extracted from the Arabic Learner Corpus (ALC)~\cite{alfaifi2012arabic} and the Arabic Learner Written Corpus (ALWC)~\cite{alfaifi2012arabic} of Arabic learners. Annotation consists of three phases: automatic preprocessing, automatic spelling corrections, and human annotations. The spelling correction was automated using the morphological analysis~\cite{habash2005arabic} and disambiguation system MADA (version 3.2)~\cite{habash2009mada+}. Annotators had to fix errors in morphology, syntax, and dialect besides spelling and punctuation. QALB-14 contains 21,396 sentences, while QALB-15 contains 1,533 sentences. 

\begin{table}[h]
\centering
\caption{QALB 2014 and QALB 2015 Corpora Statistic}
\begin{tabular}{lccc|cccc}
\hline
 & \multicolumn{3}{c|}{\textbf{QALB 2014}} & \multicolumn{4}{c}{\textbf{QALB 2015}} \\
\cline{2-8}
\textbf{Property} & \textbf{Training} & \textbf{Development} & \textbf{Testing} & \textbf{Training} & \textbf{Development} & \textbf{L1-Test} & \textbf{L2-Test} \\
\hline
Sentences     & 19,411 & 1,017 & 968 & 310 & 154 & 920 & 158 \\
Words         & 1,021,165 & 53,737 & 51,285 & 43,353 & 24,742 & 48,547 & 22,808 \\
Speaker level & Native & Native & Native & Non-native & Non-native & Native & Non-native \\
\hline
\end{tabular}
\end{table}

\section{Methodology}

\begin{figure}[ht]
    \centering
    \includegraphics[width=0.8\textwidth]{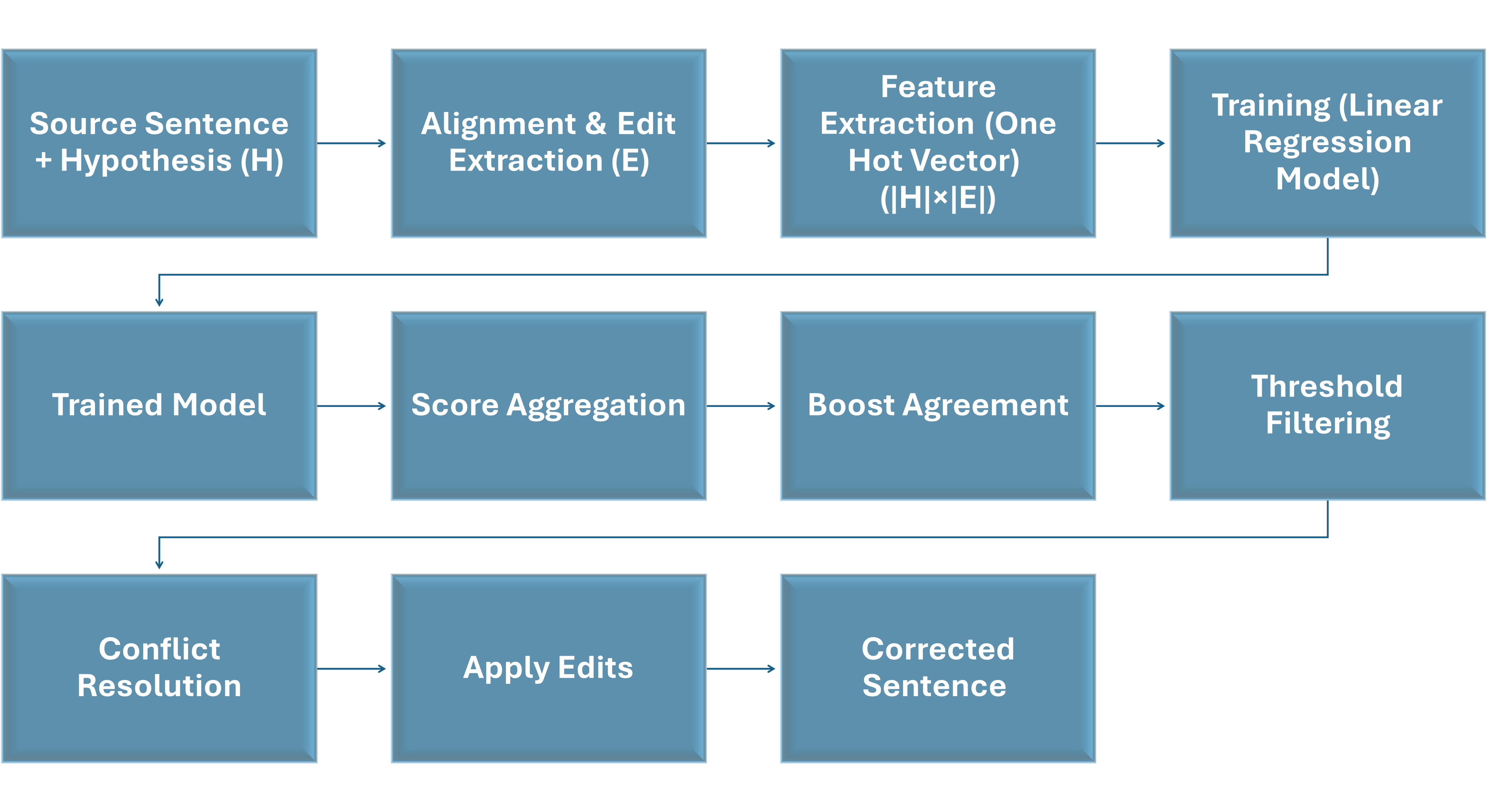}
    \caption{An illustration of the general sequence of the ArbESC+ system compilation, starting with the original sentence and hypotheses. The process then involves aligning, extracting modifications, evaluating them through the trained model, and applying the accepted modifications to generate the final corrected sentence..}
    \label{fig:pipeline}
\end{figure}

Our approach is described in this section, which combines all edits available from each hypothesis (the sentence output from each base system) and makes a decision either to keep or discard each edit to generate the final output sentence of the combined system based on the outcome of each edit. We refer to it as ARBESC+ (Arabic Enhanced Edit-based System Combination). Figure \ref{fig:pipeline} shows the operation of the ArbESC+ correction system. During the training phase, the system begins with the original sentence and hypothesis. In the next step, the two texts are aligned to extract modifications. These modifications are then converted into numerical features by converting them into a one-hot vector representation. A linear regression model uses these features to determine what modifications are correct and which are incorrect. As part of the testing phase, a new sentence with the hypothesis is presented to the system. Scores are assigned to the modifications based on the trained model. After these scores have been aggregated, they are enhanced according to the modifications that have been agreed upon. The threshold filter removes modifications with low confidence and resolves inconsistencies. Finally, the modified text is applied to the original text to produce the corrected version.

\subsection*{Algorithm 1: Arabic Enhanced Edit-Selection System Combination}
\begin{algorithm}[H]
\caption{Enhanced Edit-Selection System Combination}
\KwIn{Source sentence $s$; hypotheses $H_1,\dots,H_k$; edit types $\mathcal{T}$}
\KwOut{Corrected sentence $\hat{s}$}

\textbf{Candidate Aggregation:} collect unique edits $e=(a,b,r)$ from $H_1,\dots,H_k$.\\
\textbf{Feature Encoding:} for each $e$, build $\mathbf{x}_e\in\{0,1\}^{k\cdot|\mathcal{T}|}$ over $(H_j\times \mathcal{T})$.\\
\textbf{Scoring:} compute $p_{\mathrm{raw}}(e)=\sigma(\mathbf{w}^\top \mathbf{x}_e+b)$.\\
\textbf{Boosting:} set $p_{\mathrm{adj}}(e)=p_{\mathrm{raw}}(e)\cdot \min(1+\beta(n(e)-1),c)$.\\
\textbf{Dual-Threshold Filtering:} keep $e$ if $p_{\mathrm{raw}}(e)\ge \tau$ and $p_{\mathrm{adj}}(e)\ge \alpha\tau$.\\
\textbf{NMS on Spans:} greedily select a non-overlapping set via IoU$\le\theta$, with one insert per position.\\
\textbf{Rewrite:} apply selected edits left-to-right to obtain $\hat{s}$.\\
\Return $\hat{s}$.
\end{algorithm}
\subsection{Task Formulation}

The ArbESC+ is formulated as a binary classification problem. In our approach, we consider the GEC models as black boxes, which are then combined on the basis of the edits they propose. We derive the edits from the hypotheses generated by the base GEC models in the form of (start index, end index, correction string) sets. Additionally, each edit is also associated with an edit type derived from an automatic error annotation tool. Edit types are used as part of an edit's features. The edit type convention is based on Bryant et al.~\cite{bryant2017automatic} An edit can be one of three types: insertion (indicated by 'M:' for 'missing'), modification (indicated by 'R:' for 'replacement'), or deletion (represented by 'U:' for 'unnecessary'). 

In our method, we combine the edits from all hypotheses into a unified set, E. The edits are evaluated independently without the information of surrounding edits. As with \cite{qorib2022frustratingly}, our method relies entirely on the edit type and does not utilize the textual information. A generalized linear model is used to predict the probability that every edit from e should be retained or rejected when creating the output sentence based on edit features.
\subsection{Features}
Assume that we need to combine k hypotheses(H) from n base (component) systems. The source sentence should be a token sequence $s=(w_1,\dots,w_T)$ and $H_1,\dots,H_k$ be $k$ hypotheses. A candidate edit is a tuple $e=(a,b,r)$ where $(a,b)$ is a token span ($1\le a<b\le T$) and $r$ is the replacement string. Let $\mathcal{E}$ be the set of unique candidate edits aggregated across hypotheses. Each edit has an edit type $t \in \mathcal{T}$.

\paragraph{Feature Encoding.}
For each edit $e$, define a binary feature vector $\mathbf{x}_e \in \{0,1\}^{k\cdot|\mathcal{T}|}$:
\[
\big[\mathbf{x}_e\big]_{(j,t)}=\mathbf{1}\{\,\text{hypothesis }H_j\text{ proposes }e\text{ with type }t\,\}.
\]

In hypothesis i, $w_i$ represents e's edit type as a one-hot vector such as [0,1,0,...] if it exists, otherwise a zero vector. Consequently, the model learns whether to keep an edit entirely based on its type and which hypotheses propose it. 
\subsection{Model}
The classification model we are using is logistic regression. The logistic regression classifier trains the model with only a modest amount of training data and makes the results interpretable. Furthermore, we found that it was very effective in combining different types of base GEC systems. To calculate the likelihood that an edit is correct, we use the following formula:
\[
p_{\mathrm{raw}}(e) = \sigma(\mathbf{w}^\top \mathbf{x}_e + b),\qquad
\sigma(z)=\tfrac{1}{1+e^{-z}}.
\]
where $\sigma$ is the sigmoid function. 
Parameters $(\mathbf{w},b)$ are optimized by binary cross-entropy:
\[
\mathcal{L}(\mathbf{w},b) = -\sum_{e\in\mathcal{E}_{\text{train}}}\big[
y_e \log p_{\mathrm{raw}}(e) + (1-y_e)\log(1-p_{\mathrm{raw}}(e)) \big].
\]

\subsection{Post-processing}
Multiple hypotheses may have overlapped edits because we combine the edits from different hypotheses. 
\begin{itemize}
    \item When multiple insertion edits are proposed by different base GEC systems at the same location, they should not be applied simultaneously. In the case of multiple insertion edits at the same position (e.g., (3, 3, on) and (3, 3, in)), we consider this a conflict between multiple insertion edits.
\item If an edit's start or end index lies between another edit's start and end indices (for example, (2, 4, eaten) and (2, 3, ate)), we consider it to be an overlapping substitution conflict. 
\end{itemize}

In order to select the edits, we use a greedy strategy after obtaining the model ensemble's probabilities. In the first step, we discard any edits with raw probabilities below a predefined threshold. For the remaining edits, we multiply the raw probability by a boost factor that increases with the number of models agreeing on the same change. Our next step involves sorting the edits despondingly by their adjusted scores and iteratively selecting only those edits that don't overlap previous selections. During this stage, the output modifications are improved, and their consistency is ensured. A three-step process is involved:
\paragraph{Agreement Boosting.}
Agreement boosting increases the reliability of modifications that multiple systems support. The raw probability is multiplied by the number of systems proposing the same modification, with a cap to prevent overfitting. Two thresholds are then applied: one to the raw probability and another to the adjusted probability after boosting. This process ensures that only highly confident modifications are selected.

Let $n(e)$ be the number of hypotheses that propose $e$ (agreement count).
Define a capped boost with factor $\beta\!\ge\!0$ and cap $c\!>\!0$:
\[
\mathrm{boost}(e)=\min\big(1+\beta\cdot(n(e)-1),\,c\big),\qquad
p_{\mathrm{adj}}(e)=p_{\mathrm{raw}}(e)\cdot \mathrm{boost}(e).
\]

\paragraph{Confidence Filtering (dual thresholds).} A double filtering step is essential for selecting the most reliable edits among all candidates. We impose two complementary conditions in this step: one using the raw score, and the other using the adjusted score after applying the agreement-boosting hypothesis.

Edits evaluated by the model as statistically strong can be passed under the first condition, and the second condition adds additional assurance that the edits will remain consistent even with the support of other systems. As a result, only edits that combine the model's baseline confidence with the consensus strength are accepted.

As a result of this double filtering, we eliminate weak or unstable edits and limit the passage of edits that may be good from one perspective but lack sufficient support from an overall evaluation perspective. Consequently, a selection of highly confident edits enhances the quality of the final corrected sentence and reduces the possibility of introducing errors.
Given a main threshold $\tau\in(0,1)$ and a relaxed scale $\alpha\in(0,1]$
\[
\mathcal{C}_\tau = \{\, e \in \mathcal{E}\;\mid\;
p_{\mathrm{raw}}(e) \ge \tau \ \wedge\  p_{\mathrm{adj}}(e) \ge \alpha\tau \,\}.
\]

\paragraph{Conflict Resolution via NMS on 1D Spans.}  The Agreement Boosting stage may lead to conflicts between modifications (such as two modifications covering the same segment of text). Based on the highest scores, the Non-Maximum Suppression (NMS) algorithm selects a non-overlapping set of modifications based on the highest scores, with the additional constraint that prevents multiple insertions at the same position.
For $e_i=(a_i,b_i,r_i)$ and $e_j=(a_j,b_j,r_j)$ define 1D Intersection-over-Union:
\[
\mathrm{IoU}(e_i,e_j) = \frac{\max(0,\ \min(b_i,b_j)-\max(a_i,a_j))}
{(b_i-a_i)+(b_j-a_j)-\max(0,\ \min(b_i,b_j)-\max(a_i,a_j))}.
\]
Given $\theta \in [0,1]$, select a maximal non-overlapping set $\mathcal{G}\subseteq \mathcal{C}_\tau$ by greedy NMS:
sort $\mathcal{C}_\tau$ by $p_{\mathrm{adj}}$ (desc), then add an edit iff its IoU with all previously selected edits is $\le \theta$. Inserts ($a=b$) are additionally limited to one per position.

\paragraph{Applying Edits.} A corrected sentence is produced by combining the outputs of all three systems in left-to-right order and applying them to the source sentence. Precision and recall are improved by this structure, which is reflected in improving the $F_[0.5]$ metric for evaluating correction quality.
Sort $\mathcal{G}$ by $(a,b)$ ascending and rewrite $s$ left-to-right:
\[
\hat{s}=\mathrm{Apply}\big(s,\mathcal{G}\big).
\]

\section{Experiments}
\subsection{Implementation}
In this study, hypotheses were derived from models specifically fine-tuned for Arabic grammar correction (AraT5, ByT5, MT5), as well as AraBART+Morph+GEC QALB-2014, AraBART+Morph+GEC QALB-2015, AraBART+Morph+GEC ZEABUQ, Text editing QALB-2014, and Text editing ZEABUQ based on data from QALB-2014 and QALB-2015 training, evaluation, and testing datasets. As a result, we have 8 files for each training, evaluation, and testing set. The outputs were then converted to .m2 files using the Arabic-GEC repository in the CAMeL-Lab\footnote{\url{https://github.com/CAMeL-Lab/arabic-gec/tree/master/alignment}}. There were only three types of edits after generation: Insert, Delete, and Replace. On the other hand, QALB-2014 and QALB-2015 m2 data in the original data repository were manually annotated by a human annotator, including more types of edits, such as merge, split, and insert\_after. We manually unified all modifications in QALB-2014 and QALB-2015 m2 files into only insert, delete, and replace edits, so that edit types could be integrated across files.

Three independent experiments were conducted using ArbESC+: the first using only the QALB-2014 training data, the second using only the QALB-2015 data, and the third combining the two datasets.

The models were implemented using PyTorch's Linear module~\cite{paszke2019pytorch}, with weights optimized using the Stochastic Gradient Descent (SGD) algorithm. Loss function was Binary Cross-Entropy, with a threshold of 0.5 applied to determine modifications. Table \ref{tab:train-config} shows the model settings and parameters.

\begin{table}[t]
\centering
\caption{Training configuration and selected hyperparameters.}
\label{tab:train-config}
\begin{tabular}{ll}
\toprule
\textbf{Parameter} & \textbf{Setting} \\
\midrule
Input dimensionality & 27k \\
Learning algorithm & SGD (Stochastic Gradient Descent) \\
Loss function & Binary Cross-Entropy \\
Loss reduction mode & Mean \\
Batch size & 16 \\
Learning rate & 0.1 \\
Weight regularization & 0 (no decay) \\
Momentum factor & 0 \\
Dampening factor & 0 \\
Data shuffling (train) & Enabled \\
\bottomrule
\end{tabular}
\end{table}

\subsection{Evaluation Metrics}
We calculate precision (P), recall (R), F1, and F0.5 scores using the MaxMatch (M2) scorer~\cite{dahlmeier2012better}. Using the M2 scores, GEC systems are evaluated by comparing hypothesis edits with reference edits. In F0.5, precision is weighed twice as heavily as recall, prioritizing edit accuracy across all edits
\section{Results}
\subsection{Results of ARBESC+ in comparison to  the best single model system}
According to table \ref{tab:ensemble-comparison}, our ArbESC+ approach is compared against ensemble methods for three QALB test sets: QALB-14, QALB-15-L1, and QALB-15-L2. As shown in Table \ref{tab:qalb-dev-results}, model F0.5 scores on the QALB-14-L1 Test set approached 80\% for AraT5 and Text-edit-qalb14. However, relying on one model is limited, as each makes specific types of errors or has weaknesses in specific linguistic areas.

The performance of ensemble methods, however, was clearly improved over that of individual models. In the QALB-14-L1 Test set, the ArbESC+ model earned an F0.5 score of 82.63\%, which is an increase of about two percentage points over the best individual model. The ArbESC+ model scored 84.64 percent on the QALB-15-L1 Test set, an increase of approximately four points over the AraT5, the best individual model. This demonstrates the value of ensemble techniques, as combining inputs at this level enables the leveraging of the strengths of each model.

As a result of the complexity of the text and the variety of errors in the QALB-L2-15 Test set, ensemble techniques also provided more consistent performance across different datasets. As a result of using ArbESC+, performance improved to 65.55\%, compared to only 64.22\% for the best single model (AraT5). In addition to improving the overall average (+1.33 points), the system combination also enhances the models' ability to cope with more challenging cases and less regular data.
These results illustrate a fundamental characteristic of natural language processing tasks, particularly grammatical error correction, which is the fact that different models are often complementary rather than competitive. Some models may make mistakes in subtle grammatical structures, but others can get it correct.

In conclusion, the study confirms the importance of ensemble and system combinations in grammatical error correction tasks in Arabic, since they provide real added value that exceeds a single model in both stability and performance.

\subsection{Results of ARBESC+ in comparison to Ensemble Baseline Methods}
The ArbESC+ was compared against several common baselines in the literature, including Majority Voting, Weighted Voting, and Minimum Bayes Risk (MBR). Traditionally, these methods represent the simplest form of ensemble methods; therefore, they are commonly used as benchmarks to determine whether an ArbESC method is superior.

The baseline methods are superior to individual models, but not as effective as ArbESC+, as shown in Table \ref{tab:ensemble-comparison}. The QALB-L1-15 Test set, Weighted Voting achieved 79.56\%, ESC achieved 81.73\%, and ArbESC+ exceeded 84.64\%. In QALB-L1-14 Test set, Weighted Voting achieved 80.51\%, ESC achieved 81.72\%, and ArbESC+ exceeded 82.63\%, and in QALB-L2-15 Test set, Weighted Voting achieved 63.86\%, ESC achieved 60.59\%, and ArbESC+ exceeded 65.55\%. Even though the difference is numerically small, it means more correct corrections and fewer unnecessary corrections in grammatical error correction.

These baselines are limited by their nature of operation. With both simple and weighted voting, each sentence is treated as a whole, so the final decision is determined by which corrections received the most votes, regardless of the quality of the subtle modifications. Thus, if one model proposes a single correct modification, but the other models ignore it, simple voting may reject that modification and prefer a less accurate output simply because it was repeated more often. Due to this feature, traditional voting cannot capture subtle corrections, which are often necessary to improve the quality of Arabic text.

As for MBR, it is considered a relatively advanced baseline since it selects the output that minimizes the expected loss based on similarity to the rest of the candidates. As a result, MBR was unable to keep up with the performance of edit-level strategies. As an example, MBR achieved 81.06\% at F0.5 on the QALB-14-L1 test set, while ArbESC+ achieved 82.63\%. This difference is because MBR focuses on the sentence as a whole and lacks a mechanism to capture subtle partial edits, whereas ESC and ArbESC+ analyze each edit within a sentence, which increases the possibility of retaining correct corrections despite disagreements between models.

Across all test sets, it is evident that the ensemble baseline method provides limited quantitative improvement, whereas ArbESC+ provides more stable and significant qualitative improvements. Additionally, they improve the quality of corrections and make the system better able to handle Arabic texts with complex structures or uncommon errors.

In conclusion, simple baselines are insufficient to achieve advanced performance in correcting Arabic grammatical errors. ArbESC+ has proven to be more valuable than traditional baselines.

\subsection{Ablation Study}

Our ablation study evaluated the impact of the number of models involved in the ensemble, as well as the threshold value in the ArbESC+ mechanism. 
%% ينقص نص توضيحي عن المودل التي جمعها 

First, after testing the ensemble methods with the best three models, four models, and five models, we compared them with an ensemble method that included all nine models. The results indicate that adding models improves performance relative to one model, but this improvement does not follow a linear progression. QALB-14-L1 test set results, for instance, showed that combining the best 3 models led to 80.71\% F0.5, while combining all nine models led to 80.78\%. As a result, diversity among models leads to improved coverage and better error correction. Adding more models, however, may introduce some noise from inaccurate corrections, impairing improvements. For example, QALB-14-L1 test set results showed that combining the three best models led to 80.71\% F0.5, while combining all four models led to 79.20\%.

The ARBESC+ system relies on nine different models, the same models presented and discussed in Table \ref{tab:qalb-results} . As a result, the system does not rely on a single model alone, but rather on several models that complement each other. When applied to diverse, realistic texts, ARBESC+ can handle different types of linguistic errors better and performs more effectively

Next, we examined how changing the threshold for accepting modifications affected ArbESC+. The threshold determines how strictly the system accepts modifications proposed by different models. In order to balance precision and recall, threshold adjustment is crucial. The performance on the QALB-L2-15 test set improved from 60.59\% using traditional ESC to 65.55\% using ArbESC+ when using more conservative settings. This significant improvement (+5 points) demonstrates that the system gained from ignoring less confident modifications, retaining only those corrections with a higher degree of agreement.

In addition, all datasets do not have the same optimal threshold value. A moderate threshold (0.8) was required in the QALB-L1-15 Test, where the text was closer to formal language, and a moderate threshold (0.7) in the QALB-L1-14 Test, but a lower threshold was required to ensure stability and quality of corrections in the QALB-L2-15 Test, which had the noisiest and most diverse set of errors.

In conclusion, the performance of the system is determined by two complementary factors: the number of models involved in the ensemble and the threshold value, which determines the level of confidence required to adopt the corrections. The ArbESC+ performs significantly better than traditional baselines because it uses more models and fine-tunes the threshold, allowing for a better balance between accuracy and coverage.

\subsection{Dataset-level Analysis}
The results of the QALB-14-L1 test, the QALB-L1-15 test, and the QALB-L2-15 test were analyzed across three datasets in order to assess the performance of the models and combination methods. It is evident from the results of this study that the effect of combination varies depending on the nature of the errors and the linguistic characteristics of the texts.

The results indicate that only a limited amount of fusion strategy benefits the QALB-14-L1 Test set, which comprises texts related to the first educational level. In terms of performance, ArbESC+ achieved the best result 82.63\% F0.5, only two points behind the best single model AraT5. As these errors are more typical and repetitive, individual models will not be able to handle them efficiently since they are more common and repetitive. Consequently, fusion offers limited opportunities for improvement.

In QALB-L1-15 Test Set, the text types and errors are more diverse, which explains the differences between the individual models and the fusion methods. ArbESC+ performed 80.06\% better than AraT5 by 84.64\%. Combining the models improved both accuracy and coverage, and capturing error patterns that individuals were unable to address consistently was possible with the fusion. This study shows that fusion can be useful both for complex data sets and intermediate-level texts, which are more variable with regards to model outputs.

All models perform relatively below 65\% F0.5 on the QAL2-15 Test set, reflecting the complexity and compound errors of the test set. Despite this, fusion performed better than AraT5, showing gains of 55.90\% (AraT5) to 65.55\% (ArbESC+). Fusion's effectiveness is evident in this significant improvement (+9.6 points), especially when individual models fall short of providing sufficient coverage. Because of the variety of errors in this set, every model captures different parts of the data. As a result, combining these parts via an intelligent mechanism like ArbESC+ produces significantly better results than each component alone.

Based on this analysis, we find that fusion gains depend on the complexity of the dataset. It is limited to simpler error sets (QALB-14-L1 Test) since most errors can be handled by individual models. As more complex sets (QALB-L2-15 Test) are tested, fusion is more pronounced, compensating for the lack of coverage of individual models and delivering more stable and higher performance. Thus, fusion improves not only overall performance but also the system's ability to adapt to different data sets and error levels.
\begin{table*}[ht]
\centering
\renewcommand{\arraystretch}{1.15}
\setlength{\tabcolsep}{4pt}
\resizebox{\textwidth}{!}{%
\begin{tabular}{l|cccc|cccc|cccc}
\toprule
\multirow{2}{*}{\textbf{Ensemble Method}} & \multicolumn{4}{c|}{\textbf{QALB-14-L1 Test}} & \multicolumn{4}{c|}{\textbf{QALB-L1-15 Test}} & \multicolumn{4}{c}{\textbf{QALB-L2-15 Test}} \\
\cmidrule(lr){2-5}\cmidrule(lr){6-9}\cmidrule(lr){10-13}
 & P & R & F1 & F0.5 & P & R & F1 & F0.5 & P & R & F1 & F0.5 \\
\midrule
Weighted MV (sentence-level) & 82.56\% & \textbf{73.25\%} & \textbf{77.63\%} & 80.51\% & 79.65\% & \textbf{79.47\%} & \textbf{79.56\%} & 79.61\% & 71.39\% & 44.91\% & 55.13\% & 63.86\% \\
Edit-level MV (best 3 models) & 86.13\% & 64.48\% & 73.75\% & 80.71\% & 83.08\% & 70.87\% & 76.49\% & 80.31\% & 73.89\% & 37.22\% & 49.50\% & 61.73\% \\
Edit-level MV (best 4 models) & 88.91\% & 55.13\% & 68.06\% & 79.20\% & 86.14\% & 60.82\% & 71.30\% & 79.52\% & \textbf{79.29\%} & 26.78\% & 40.04\% & 56.95\% \\
Edit-level MV (best 5 models) & 87.09\% & 62.60\% & 72.84\% & 80.77\% & 83.83\% & 70.16\% & 76.39\% & 80.69\% & 77.79\% & 32.56\% & 45.90\% & 60.87\% \\
Edit-level MV(All 9 models) & 83.66\% & 71.02\% & 76.82\% & 80.78\% & 81.09\% & 76.93\% & 78.96\% & 80.22\% & 72.56\% & 41.30\% & 52.64\% & 63.02\% \\
MBR  & 84.52\% & 69.65\% & 76.37\% & 81.06\% & 82.45\% & 76.33\% & 79.27\% & 81.15\% & 71.26\% & 39.73\% & 51.02\% & 61.50\% \\
ESC & 85.47\% & 69.51\% & 76.67\% & 81.72\% & 83.98\% & 73.83\% & 78.58\% & 81.73\% & 65.55\% & \textbf{46.52\%} & 54.42\% & 60.59\% \\
ArbESC+ (QALB-14) & 87.12\% & 66.83\% & 75.64\% & 82.13\% & 89.27\% & 66.28\% & 76.08\% & 83.48\% & 74.46\% & 42.80\% & 54.36\% & 64.86\% \\
ArbESC+ (QALB-15) & 88.35\% & 62.26\% & 73.05\% & 81.52\% & 90.55\% & 63.85\% & 74.90\% & 83.56\% & 70.40\% & 44.44\% & 54.49\% & 63.04\% \\
ArbESC+ & \textbf{90.33}\% & 61.61\% &  73.25\% & \textbf{82.63\%} & \textbf{92.18\%} & 63.79\% & 75.40\% & \textbf{84.64\%} & 73.47\% & 45.52\% & \textbf{56.27\%} & \textbf{65.55\%} \\
\bottomrule
\end{tabular}%
}

\caption{Comparison of ensemble strategies across QALB test sets.}
\label{tab:ensemble-comparison}
\end{table*}

\begin{table*}[ht]
\centering
\renewcommand{\arraystretch}{1.15}
\setlength{\tabcolsep}{4pt}
\resizebox{\textwidth}{!}{%F
\begin{tabular}{l|cccc|cccc}
\toprule
\multirow{2}{*}{\textbf{Ensemble Method}} & \multicolumn{4}{c|}{\textbf{QALB-14-L1 Dev}} & \multicolumn{4}{c}{\textbf{QALB-L2-15 Dev}} \\
\cmidrule(lr){2-5}\cmidrule(lr){6-9}
 & P & R & F1 & F0.5 & P & R & F1 & F0.5 \\
\midrule
Weighted MV (all nine models) & 81.68\% & \textbf{74.06\%} & \textbf{77.68\%} & 80.03\% & 70.83\% & 44.78\% & 54.87\% & 63.45\% \\
MV (best 3 models) & 84.80\% & 65.51\% & 73.92\% & 80.08\% & 72.79\% & 36.72\% & 48.82\% & 60.84\% \\
MV (best 4 models) & 87.49\% & 55.41\% & 67.85\% & 78.41\% & 72.26\% & 41.40\% & 52.64\% & 62.88\% \\
MV (best 5 models) & 85.86\% & 63.25\% & 72.84\% & 80.13\% & 72.19\% & \textbf{67.66\%} & \textbf{55.52\%} & 64.45\% \\
MV (all 9 models) & 82.62\% & 71.53\% & 76.67\% & 80.13\% & 72.27\% & 41.38\% & 52.63\% & 62.88\% \\
MBR & 83.45\% & 70.53\% & 76.45\% & 80.50\% & 71.17\% & 40.55\% & 51.66\% & 61.83\% \\
ESC & 84.93\% & 70.73\% & 77.18\% & 81.65\% & 69.15\% & 46.04\% & 55.28\% & 62.84\% \\
ArbESC+ (QALB-14 tuned) & 86.56\% & 68.22\% & 76.31\% & 82.15\% & 74.82\% & 43.80\% & 55.25\% & 65.53\% \\
ArbESC+ (QALB-15 tuned) & \textbf{88.35\%} & 58.06\% & 70.75\% & 81.42\% & \textbf{79.79\%} & 34.98\% & 48.64\% & 63.52\% \\
ArbESC+   & 89.85\% & 59.16\% & 71.73\% & \textbf{82.78\%} & 76.63\% & 41.81\% & 54.10\% & \textbf{65.69\%} \\
\bottomrule
\end{tabular}%
}
\caption{Comparison of ensemble strategies across QALB Dev sets.}
\label{tab:ensemble-comparison-2}
\end{table*}

\subsubsection{Comparison with Existing GEC Frameworks}

 ArbESC+ performed better than previous models when it was used to correct linguistic errors on both the QALB-2014 and QALB-2015 corpora, outperforming all previous models in terms of F0.5 score. Table \ref{tab:comparison_qalb_2014_2015} shows that the model achieved the highest values across most metrics, with an F1 score of 73.25\% and an F0.5 of 82.63\% for QALB-2014 and 75.40\% and 84.64\% for QALB-2015.

ArbESC+'s superiority can be attributed to its dual architecture, which integrates error correction for multiple single models into an integrated framework, as well as intelligent re-ranking mechanisms that analyze subtle errors and select the most effective correction. As a result of the edit selection mechanism, the model also became more accurate at handling complex grammatical and spelling errors, especially those requiring precise discrimination between closely related linguistic alternatives.

Our proposed model performed significantly better than previous models. As a result, erroneous corrections are reduced while the original meaning is preserved. ArbESC+ emphasizes high accuracy combined with good generalization, providing a promising framework for correcting Arabic texts with a level of accuracy that is comparable to human.
\begin{table*}[ht!]
\centering
\caption{Comparison of our model correction performance against prior and recent work on the QALB-2014 and QALB-2015 datasets.}
\small
\begin{tabular}{lcccccccc}
\toprule
\textbf{Model / Study} & \multicolumn{4}{c}{\textbf{QALB-2014}} & \multicolumn{4}{c}{\textbf{QALB-2015}} \\
\cmidrule(lr){2-5} \cmidrule(lr){6-9}
& \textbf{Prec.} & \textbf{Rec.} & \textbf{F$_1$} & \textbf{F$_{0.5}$} & \textbf{Prec.} & \textbf{Rec.} & \textbf{F$_1$} & \textbf{F$_{0.5}$} \\
\midrule
Rozovskaya et al. (2014)\cite{rozovskaya2014qalb} & 72.22 & 62.79 & 67.18 & 70.12 & - & - & - & - \\
Nawar (2015)\cite{nawar2015arabic} & - & - & - & - & 88.85 & 61.76 & 72.87 & 81.69 \\
Aiman et al. (2021)\cite{aiman2021deep} & - & - & - & - & 80.23 & 63.59 & 70.91 & 76.21 \\
Pajak and Pajak (2022)\cite{pajak2022neural} & - & - & - & - & 75.91 & 62.65 & 68.64 & 72.87 \\
Solyman et al. (2022)\cite{solyman2022hybrid} & 79.06 & 65.79 & 71.82 & 75.97 & 78.36 & 70.43 & 74.18 & 76.63 \\
Solyman et al. (2023)\cite{solyman2023arat5} & 78.66 & 65.04 & 71.03 & 75.47 & 77.68 & 69.78 & 73.52 & 75.47 \\
Kwon et al. (2023) / GPT-4 (5-shot)\cite{kwon2023gpt4} & 69.46 & 61.96 & 65.49 & 67.82 & 52.33 & 47.57 & 49.83 & 54.10 \\
Mahmoud et al. (2024)a\cite{mahmoud2024hybrid} & 72.8 & \textbf{71.2} & 73.9 & 75.6 & 74.0 & \textbf{75.4} & 74.7 & 74.3 \\
Mahmoud et al. (2024)b\cite{mahmoud2024t5} & 80.14 & 67.2 & 73.1 & 77.16 & 81.35 & 70.43 & 75.49 & 78.89 \\
Alrehili et al. (2025)\cite{alrehili2025synthetic} & 81.88 & 70.68 & 75.87 & 79.36 & 56.77 & 72.10 & 62.99 & 68.57 \\
Habash (2025)\cite{hefny2025textedit} & 89.7 & 60.2 & 72.0 & 81.7 & 88.3 & 66.7 & 76.0 & 82.9 \\
\midrule
\textbf{ArbESC+ (QALB-14)} & 87.12 & 66.83 & 75.64 & 82.13 & 89.27 & 66.28 & 76.08 & 83.48 \\
\textbf{ArbESC+ (QALB-15)} & 88.35 & 62.26 & 73.05 & 81.52 & 90.55 & 63.85 & 74.90 & {83.56} \\
\textbf{ArbESC+ (Overall)} &  90.33\% & 61.61\% &  73.25\% & \textbf{82.63\%} & \textbf{92.18\%} & 63.79\% & 75.40\% & \textbf{84.64}\% \\
\bottomrule
\end{tabular}
\label{tab:comparison_qalb_2014_2015}
\end{table*}

\section{Error-types Analysis}
We used the ARETA tool~\cite{belkebir2021automatic} to analyze the grammatical errors that remain in text after they have been corrected using a single model system and our proposed model ARBESC+. In ARETA, each source sentence is linked to a corrected version. Edits are automatically tagged based on ARETA's taxonomy. In this evaluation, the accuracy of the correction is assessed by how effectively each model corrected the ARETA gold errors. 

Nine models were evaluated in the same ARETA configuration:
M1: AraT5, M2: ByT5, M3: mT5, M4: AraBART+Morph+GEC13 (2014), M5: AraBART+Morph+GEC13 (2015), M6: AraBART+Morph+GEC (ZEABUQ), M7: Text-Edit-14, M8: Text-Edit-ZEABUQ, M9: AraBART

Analyses were conducted on the following data sets: QALB14-L1, QALB15-L1, and QALB15-L2. This structure enabled us to assess the robustness of our model and observe domain shifts between native Arabic and learner Arabic.
\subsection{Single Model Analysis}
\subsubsection{Analysis of Error-Class Performance}
As shown in Table \ref{tab:areta_qalb_comparison}, the models in the QALB14-L1, QALB15-L1, and QALB15-L2 test set vary across error categories. Each model is superior depending on its correction strategy and the nature of the category. The results show that the M8 model (Text-Edit-ZEABUQ) achieves the best results across most corpora in punctuation. Due to its limited-positional-editing methodology, the corrections are restricted to clear, small portions without rewriting the whole sentence. As punctuation errors often require a single character to be added or removed, the approach is aligned with the nature of punctuation errors to reduce the likelihood of overcorrection. In addition, the model benefits from corpus-based training using different distributions of punctuation errors, enhancing its sensitivity to Arabic punctuation patterns. There are two specialized text editing models: one focused on punctuation correction, and the other on word correction edits. It improves performance since the former deals with a task that is similar to predicting the placement or removal of punctuation marks, a relatively simple categorical process, while the latter deals with morphological/syntactic agreements, spelling, inflection, and word reordering.

Thus, the superiority of the M8 model in the “punctuation” category in the table \ref{tab:areta_qalb_comparison} can be attributed to its actual implementation of a text editing model, having a dedicated punctuation model and a dedicated word correction model, which reduces mutual interference between tasks with different natures and improves performance.

The AraT5(M1) model outperformed in all test sets with regard to orthographic errors. The model has a high level of accuracy due to its pre-trained model on extensive Arabic data and ability to represent affixes, alif and hamza changes, and vowel extensions across texts, the model has a high level of accuracy. Because of these features, it can address subtle orthographic variations common in Arabic texts, especially those caused by differences between standard and colloquial usage.

The AraBART+Morph+GEC13(M5) dominates in terms of morphological errors. A fine-tuning process incorporating morphological analysis ensures this superiority. Morphological features should be represented to ensure the correct word form is selected and that morphological units within a sentence correspond. Due to this model, the word is not only corrected on its textual surface, but also grammatically corrected within its syntactic structure.

On the other hand, semantics is more complex because it relies more on context than surface form. This results in a better performance for native (L1) data when using ByT5 (M2). Due to its byte-level operation, it can be sensitive to subtle character changes that could lead to a shift in meaning, especially with negations, pronouns, and clause connectives. When comparing L2 data with AraT5(M1), the superiority shifted to M1, which indicates that linguistic knowledge is necessary to correct learner errors, rather than just detecting positional deviations.

Finally, limited-editing models such as M8 can be particularly beneficial for the syntax category, since they limit unnecessary changes in word order. The M1 sequence model performs better on learner data because a small positional change requires more syntactic reconstruction.

In general, AraT5 was the most stable and adaptable across all test conditions, whereas AraBART-based variants contributed specialized gains in punctuation and morphology. The combined generalization of AraT5 and the morphological refinement of AraBART would result in a GEC system that is more comprehensive and linguistically balanced.

Table \ref{tab:areta_dev_comparison} illustrates how errors affect model performance on the development datasets. While pre-trained Arabic models such as AraT5 perform best in spelling and merging, AraBART+Morph excels in morphological errors, which require the analysis of grammatical features and their relationships with one another. Text-Edit-ZEABUQ, however, performs best in the punctuation class because it separates punctuation and word correction from over-editing, allowing original sentence structure to be preserved. The data from both natives (L1) and learners (L2) demonstrate clear differences, with the latter requiring more structural modifications, enhancing the performance of sequential models.
\begin{table*}[ht]
\centering
\renewcommand{\arraystretch}{1.1}
\setlength{\tabcolsep}{4pt}
\caption{ARETA-based Class-Level Correction Accuracy across QALB14-L1, QALB15-L1, and QALB15-L2 Test Sets.
Model1 = AraT5, Model2 = ByT5, Model3 = mT5, Model4 = AraBART+Morph+GEC13 (2014),
Model5 = AraBART+Morph+GEC13 (2015), Model6 = AraBART+Morph+GEC (ZEABUQ),
Model7 = Text-Edit-14, Model8 = Text-Edit-ZEABUQ, Model9 = AraBART.}
\label{tab:areta_qalb_comparison}
\begin{tabular}{llccccccccccc}
\hline
\textbf{Dataset} & \textbf{Class} & \textbf{Actual Errors} &
\textbf{M1} & \textbf{M2} & \textbf{M3} & \textbf{M4} & \textbf{M5} & \textbf{M6} & \textbf{M7} & \textbf{M8} & \textbf{M9} & \textbf{Best}\\
\hline
\multirow{7}{*}{QALB14-L1}
& Merge          & 809  & \textbf{91.97} & 70.83 & 89.74 & 88.13 & 87.39 & 87.39 & 88.13 & 89.62 & 89.49 & M1\\
& Morphological  & 95   & 43.16 & 37.89 & 37.89 & 44.21 & \textbf{52.63} & 47.37 & 34.73 & 44.21 & 37.89 & M5\\
& Orthographic   & 7322 & \textbf{85.30} & 84.13 & 80.90 & 79.53 & 81.36 & 80.66 & 79.72 & 81.51 & 81.35 & M1\\
& Punctuation    & 6853 & 63.99 & 62.42 & 58.20 & 57.13 & 61.95 & 60.33 & 63.64 & \textbf{71.25} & 60.81 & M8\\
& Semantic       & 234  & 71.80 & \textbf{75.22} & 71.36 & 69.23 & 67.09 & 67.95 & 57.69 & 58.55 & 63.68 & M2\\
& Split          & 424  & \textbf{100.0} & \textbf{100.0} & \textbf{100.0} & \textbf{100.0} & \textbf{100.0} & \textbf{100.0} & \textbf{100.0} & \textbf{100.0} & \textbf{100.0} & All\\
& Syntax         & 839  & 35.76 & 33.25 & 30.15 & 30.75 & 33.73 & 32.66 & 33.37 & \textbf{36.47} & 30.16 & M8\\
\hline
\multirow{7}{*}{QALB15-L1}
& Merge          & 676  & \textbf{92.75} & 66.72 & 91.12 & 89.35 & 90.09 & 90.38 & 91.42 & 91.12 & 92.75 & M1\\
& Morphological  & 89   & \textbf{43.82} & \textbf{43.82} & 29.21 & 34.83 & 42.69 & \textbf{43.82} & 30.34 & 40.45 & 43.82 & M1/M2/M6\\
& Orthographic   & 2549 & \textbf{67.05} & 61.55 & 53.24 & 55.36 & 54.61 & 53.27 & 50.96 & 57.16 & 67.05 & M1\\
& Punctuation    & 4210 & 66.29 & 61.30 & 57.27 & 56.77 & 62.78 & 60.53 & 67.39 & \textbf{73.31} & 66.29 & M8\\
& Semantic       & 145  & 67.58 & \textbf{70.34} & 64.82 & 53.80 & 57.24 & 55.17 & 56.55 & 62.07 & 67.58 & M2\\
& Split          & 397  & \textbf{100.0} & \textbf{100.0} & \textbf{100.0} & \textbf{100.0} & \textbf{100.0} & \textbf{100.0} & \textbf{100.0} & \textbf{100.0} & \textbf{100.0} & All\\
& Syntax         & 679  & 40.65 & 42.27 & 29.60 & 35.79 & 36.82 & 36.38 & 43.01 & \textbf{44.92} & 40.65 & M8\\
\hline
\multirow{7}{*}{QALB15-L2}
& Merge          & 198  & 63.13 & 61.11 & 61.11 & \textbf{75.25} & 55.56 & 56.06 & 60.61 & 53.54 & 58.08 & M4\\
& Morphological  & 299  & 38.46 & 27.76 & 27.42 & 22.41 & \textbf{42.14} & 39.46 & 13.05 & 24.08 & 33.11 & M5\\
& Orthographic   & 1610 & \textbf{44.04} & 30.81 & 33.79 & 24.53 & 40.93 & 41.49 & 24.60 & 32.05 & 40.62 & M1\\
& Punctuation    & 2140 & 47.57 & 48.88 & 46.12 & 38.46 & \textbf{51.12} & 49.11 & 38.65 & 21.69 & 48.46 & M5\\
& Semantic       & 441  & \textbf{46.04} & 35.83 & 39.68 & 26.99 & 45.58 & 44.22 & 27.43 & 32.20 & 40.81 & M1\\
& Split          & 34   & \textbf{100.0} & \textbf{100.0} & \textbf{100.0} & \textbf{100.0} & \textbf{100.0} & \textbf{100.0} & \textbf{100.0} & \textbf{100.0} & \textbf{100.0} & All\\
& Syntax         & 1600 & \textbf{38.81} & 29.94 & 28.44 & 18.31 & 35.13 & 35.19 & 15.13 & 21.69 & 32.31 & M1\\
\hline
\end{tabular}
\end{table*}

\begin{table*}[ht]
\centering
\renewcommand{\arraystretch}{1.1}
\setlength{\tabcolsep}{4pt}
\caption{ARETA-based class-level correction accuracy on development sets. Actual Errors is ARETA+ Count, the number of gold errors from source to reference alignment. Model1 AraT5, Model2 ByT5, Model3 mT5, Model4 AraBART+Morph+GEC13 (2014), Model5 AraBART+Morph+GEC13 (2015), Model6 AraBART+Morph+GEC (ZEABUQ), Model7 Text-Edit-14, Model8 Text-Edit-ZEABUQ, Model9 AraBART.}

\label{tab:areta_dev_comparison}
\begin{tabular}{llccccccccccc}
\hline
\textbf{Dataset} & \textbf{Class} & \textbf{Actual Errors} &
\textbf{M1} & \textbf{M2} & \textbf{M3} & \textbf{M4} & \textbf{M5} & \textbf{M6} & \textbf{M7} & \textbf{M8} & \textbf{M9} & \textbf{Best}\\
\hline
\multirow{7}{*}{QALB14-L1 Dev}
& Merge          & 805  & \textbf{93.04} & 69.57 & 89.94 & 90.19 & 89.57 & 89.32 & 88.45 & 90.06 & 91.06 & M1\\
& Morphological  & 110  & 49.09 & 42.73 & 34.55 & \textbf{50.00} & 43.64 & 49.09 & 31.82 & 41.82 & 46.36 & M4\\
& Orthographic   & 3142 & \textbf{66.90} & 61.65 & 55.16 & 57.04 & 56.56 & 54.58 & 52.13 & 56.81 & 57.96 & M1\\
& Punctuation    & 6617 & 61.87 & 60.51 & 57.09 & 55.74 & 59.08 & 57.09 & 62.46 & \textbf{69.14} & 60.46 & M8\\
& Semantic       & 185  & \textbf{66.49} & 61.62 & 61.08 & 58.92 & 51.89 & 58.92 & 50.81 & 58.38 & 53.52 & M1\\
& Split          & 422  & \textbf{100.0} & \textbf{100.0} & \textbf{100.0} & \textbf{100.0} & \textbf{100.0} & \textbf{100.0} & \textbf{100.0} & \textbf{100.0} & \textbf{100.0} & All\\
& Syntax         & 896  & 39.85 & \textbf{43.08} & 28.34 & 36.38 & 33.48 & 31.81 & 38.84 & 41.29 & 32.70 & M2\\
\hline
\multirow{7}{*}{QALB15-L2 Dev}
& Merge          & 210  & 51.90 & 17.14 & 58.10 & \textbf{59.52} & 44.29 & 46.19 & 30.00 & 21.90 & 55.24 & M4\\
& Morphological  & 371  & 34.50 & 30.46 & 22.10 & 17.79 & \textbf{44.47} & 42.05 & 14.82 & 22.65 & 33.15 & M5\\
& Orthographic   & 1760 & \textbf{48.81} & 33.81 & 37.27 & 25.34 & 38.86 & 37.10 & 27.22 & 32.73 & 37.95 & M1\\
& Punctuation    & 2267 & 49.67 & 51.43 & 51.08 & 39.04 & 53.16 & 46.40 & 40.72 & 21.09 & \textbf{53.51} & M9\\
& Semantic       & 504  & 35.91 & 33.14 & 31.95 & 22.02 & \textbf{41.27} & 39.09 & 19.84 & 25.60 & 35.51 & M5\\
& Split          & 42   & \textbf{100.0} & \textbf{100.0} & \textbf{100.0} & \textbf{100.0} & \textbf{100.0} & \textbf{100.0} & \textbf{100.0} & \textbf{100.0} & \textbf{100.0} & All\\
& Syntax         & 1785 & \textbf{39.83} & 31.32 & 32.38 & 19.50 & 36.75 & 35.74 & 16.36 & 22.52 & 33.22 & M1\\
\hline
\end{tabular}
\end{table*}
\subsubsection{Analysis of Error-Type Performance}
Table \ref{tab:areta_error_types_full}, it compares the performance of the different models in correcting the ARETA errors in the QALB2014 and QALB2015 (L1 and L2) datasets. The results indicate that the models are capable of handling a wide range of morphological, orthographic, and syntactic errors, and that there are performance differences between native (L1) and non-native (L2) Arabic learners.

Generally, the models achieved high performance when predicting recurring typographic errors such as word mergers (MG), word separations (SP), and punctuation marks (PT, PM, PC). Due to their simplicity and syntactic clarity, as well as their frequency in the training data, these patterns are easy to detect by most models.

QALB14 achieved high scores in the OH class for complex spelling errors such as errors with hamza (OH), alif-ya' (OA), and alif-ya' (OW). As compared to the L2 data set, however, the L2 data set showed a decline of 37 percent. As Arabic learners as a second language use alif and hamza differently from native speakers as a second language, the model is more challenging.

However, for certain error tags (MI, MT, OM, XG, XN), the model's results ranged from 35 to 61 percent. It is difficult to discern these errors merely from a superficial reading of the text because of the complexity of the Arabic morphological system.

A semantic challenge appears to arise from word substitution (OR) and word choice (SW) categories, which require an understanding of the context in which they are used to select the best alternative. Although some models achieved good results up to 100\% in some cases, the large variance between datasets indicates how sensitive these categories are to the nature of the data and learners' different writing styles.

It is evident that these models are unable to process subtle grammatical relationships, such as definiteness and indefiniteness (XF) and agreement in number and gender (XG, XN). In this scenario, the training data will likely underrepresent these errors, and the context of the whole sentence will need to be processed.

For categories with limited patterns or fixed rules, such as disjunction (SP), extra addition (XT), and extra punctuation (PT), the models performed optimally on all datasets (100%).

Finally, model performance is significantly lower in the L2 data set than in the L1 and QALB-14 datasets. This indicates poor generalization when dealing with Arabic learners' more diverse error patterns. To address deep and semantic errors, models must be augmented with data from L2 learners or incorporate interpretive linguistic components.
\begin{table*}[ht]
\centering
\renewcommand{\arraystretch}{1.05}
\setlength{\tabcolsep}{3pt}
\caption{Best correction accuracy per ARETA+ error type across QALB14, QALB15-L1, and QALB15-L2 test sets. Values represent the highest correction accuracy (\%) and the model achieving that score. Models: M1 AraT5, M2 ByT5, M3 mT5, M4 AraBART+Morph+GEC13 (2014), M5 AraBART+Morph+GEC13 (2015), M6 AraBART+Morph+GEC (ZEABUQ), M7 Text-Edit-14, M8 Text-Edit-ZEABUQ, M9 AraBART.}
\label{tab:areta_error_types_full}
\small
\begin{tabular}{llccccccc}
\hline
\textbf{Tag} & \textbf{Error Description} & \textbf{QALB14 (\%)} & \textbf{Best (14)} & \textbf{QALB15-L1 (\%)} & \textbf{Best (L1)} & \textbf{QALB15-L2 (\%)} & \textbf{Best (L2)} \\
\hline
MG & Merge errors & 91.97 & M1 & 92.75 & M9 & 75.25 & M1 \\
MI & Morphological inflection & 51.72 & M5 & 42.68 & M5 & 43.21 & M5 \\
MT & Morphological transfer & 62.50 & M1 & 50.00 & M1 & 35.19 & M1 \\
OA & Alif–Ya confusion & 83.70 & M7 & 58.97 & M7 & 57.58 & M7 \\
OC & Character order & 100.00 & M1 & 100.00 & All & 33.33 & M8 \\
OD & Deletion errors & 66.17 & M2 & 68.42 & M9 & 55.22 & M9 \\
OG & Long–short vowel confusion & 100.00 & All & 100.00 & All & 100.00 & All\\
OH & Hamza and orthographic errors & 91.73 & M1 & 73.09 & M1 & 37.95 & M1 \\
OM & Morphological agreement & 57.69 & M1 & 61.52 & M9 & 42.94 & M1 \\
OR & Word replacement & 54.87 & M5 & 61.13 & M5 & 59.58 & M5 \\
OT & Ta–Ha confusion & 95.92 & M1 & 69.60 & M8 & 84.38 & M1 \\
OW & Alif Fariqa confusion & 85.98 & M1 & 85.71 & M1 & 55.56 & M1 \\
PC & Punctuation confusion & 100.00 & All & 100.00 & All & 10.91 & M4 \\
PM & Missing punctuation & 79.19 & M8 & 81.97 & M8 & 64.17 & M4 \\
SP & Word split & 100.00 & All & 100.00 & All & 100.00 & All \\
SW & Word selection error & 100.00 & All & 69.93 & M2 & 47.26 & M2 \\
XC & Case error & 100.00 & M1 & 59.74 & M8 & 54.34 & M8 \\
XF & Definiteness & 36.67 & M1 & 43.59 & M1/M6/M9 & 40.64 & M5 \\
XG & Gender agreement & 35.44 & M6 & 56.86 & M6 & 59.59 & M6 \\
XN & Number agreement & 46.23 & M5 & 53.19 & M5 & 50.44 & M5 \\
XM & Missing word & 41.46 & M2 & 100.00 & All & 29.67 & M3 \\
XT & Unnecessary word & 100.00 &  All & 100.00 & All & 100.00 & All \\
SF & Conjunction use & 100.00 & M1 & 100.00 & M2 & 27.27 & M2 \\
PT & Unnecessary Punctuation & 100.00 &  All & 100.00 & All & 100.00 &  All  \\
\hline
\end{tabular}
\end{table*}
\subsubsection{Over-Correction Analysis}
The overcounting of grammatically correct text is caused by models introducing unnecessary edits. As a result, the models produce significantly more raw edits than the number of correct edits adopted after review in the categories of punctuation (PT and PC), extra words (XT), and missing words (XM). Table \ref{tab:overcount_summary} presents a summary of the most affected categories in the QALB14 and QALB15 datasets (L1 and L2).
\begin{table}[!ht]
\centering
\small
\caption{Over-Correction Summary for Frequent Tags (QALB14 \& QALB15, L1/L2)}
\label{tab:overcount_summary}
\begin{tabular}{lcccc}
\hline
\textbf{Tag} & \textbf{Raw} & \textbf{Fixed} & \textbf{OCR (\%)} & \textbf{Note} \\
\hline
PT (Unnec. punct.) & 1100--1500 & 1--13 & $>$99 & Triggered by spacing/punctuation \\
PC (Punct. conf.)  & 700--1100  & 440--680 & 40--60 & Duplicate tags in same span \\
XT (Unnec. word)   & 180--2000  & 17--46  & $>$95 & Over-splitting short text \\
XM (Miss. word)    & 90--170    & 87--123 & 30--50 & Mixed delete/punct. overlap \\
\hline
\end{tabular}
\end{table}

Some categories have overcounting ratios (OCRs) greater than 95\%, particularly punctuation and separation. As a result, the models correct linguistic errors caused by slight discrepancies in spaces or punctuation. Due to this behavior, there is an inflated number of edits without any benefit to the text quality or accuracy of corrections.

 The models rely more on surface patterns than on linguistic context, learning formal relationships. Also, compound edits, such as combining a word deletion with a punctuation edit, are fragmented into multiple tags, which doubles the number of recorded edits. Furthermore, overstimulation of the PT and PC categories is caused by low decision thresholds for some tags. As a result of learner diversity in the L2 data set, whose writing styles and punctuation usage differ from standard Arabic, overstimulation is more likely to occur.

Therefore, automated correction systems should be evaluated both quantitatively and qualitatively. Overcounting analysis reveals hidden behaviors in models, especially in Arabic where spelling and syntactic errors are common, whereas precision and recall provide general indicators. Consequently, addressing this phenomenon is essential to improving model reliability and ensuring more accurate and balanced corrections.

\subsection{ArbESC+ Error Type Analysis}

In the table \ref{tab:areta_class_comparison}, the performance of ARBESC+ is shown for each test set based on the ARETA classification: QALB-2014, QALB-2015 (L1), and QALB-2015 (L2). ARBESC+ is remarkably close to the performance of the best individual model in each category when compared to the detailed results of the nine individual models in Table \ref{tab:areta_qalb_comparison}, and sometimes even outperforms it in some categories, especially in the Merge and Orthographic categories of the QALB-2014 set.

Despite the differences in text characteristics between QALB-L1 and QALB-L2, ARBESC+ maintains a stable level of performance among all three datasets. ARBESC+, their performance declines to varying degrees. However, ARBESC+ maintains a more balanced correction level, indicating that ARBESC+'s fusion mechanism not only reduces performance variability among datasets but also improves the system's ability to deal with error patterns that differ between levels of learners.

Additionally, ARBESC+ maintained this level of correction without deterioration across datasets, despite the Split category achieving a 100% correction rate. ARBESC+ achieved a stable average performance without a sharp decline in categories such as Morphology and Syntax, demonstrating the ability of the fusion mechanism to mitigate individual weaknesses.

Accordingly, ARBESC+ does not aim to outperform any single model numerically, but rather to provide consistent support across different error classes and thus be more suitable for use in a variety of text contexts than relying solely on one model.

\begin{table*}[ht!]
\centering
\caption{ArbESC+ performance (\%) across major ARETA error classes for QALB-2014, QALB-2015 (L1), and QALB-2015 (L2). Values represent the percentage of correctly corrected errors per class.}
\small
\begin{tabular}{lccc}
\toprule
\textbf{Error Class} & \textbf{QALB-2014 (\%)} & \textbf{QALB-2015 L1 (\%)} & \textbf{QALB-2015 L2 (\%)} \\
\midrule
Merge           & 92.21 & 84.32 & 76.16 \\
Morphological   & 46.84 & 43.59 & 41.77 \\
Orthographic    & 83.12 & 70.04 & 48.67 \\
Punctuation     & 67.48 & 65.22 & 54.34 \\
Semantic        & 67.27 & 70.31 & 45.82 \\
Split           & 100.0 & 100.0 & 100.0 \\
Syntax          & 35.54 & 43.43 & 38.75 \\
\bottomrule
\end{tabular}
\label{tab:areta_class_comparison}
\end{table*}

In Table \ref{tab:ArbESC_per_tag_percentages}, the subtle variations in ArbESC+ performance across the different QALB corpora within the ARETA system are shown. There are three main types of behavior to be observed:

First, there are the tag corpora for which the model consistently performs well, such as PT (punctuation), OG (vowel lengthening/shortening), and MG (word blending). As a consequence, ArbESC+'s mechanism for identifying modification positions remains effective even when transitioning from native to learner texts based on the patterns represented by these tags. It is more difficult to correct these errors using deep linguistic interpretation due to their relative independence from context.

Second, tags decline sharply when transitioning from QALB-2014 and QALB-2015 (L1) to QALB-2015 (L2), particularly SF (word order within phrase), OC (letter order within word), and OH (hamza errors). Since learners' errors (L2) do not follow standard writing patterns, they are characterized by a higher degree of randomness and syntactic variability, limiting the model's ability to predict correct corrections based on the patterns it has learned during training. As a result of this gap, L2 error correction models using contextual reconstruction instead of only positional processing are clearly needed.

Third, some tags indicate that performance in L2 is higher than in L1 in some cases, such as XG (changes associated with semantic morphemes) and XN (manipulation of negation/articles). As a result of this, the model is able to capture some errors made by learners according to a simplified grammatical logic that can be compared to errors made by native speakers.

In addition, it appears that fully structural tags, such as SP and XT, which relate to sentences or syntactic units, remain at 100\% across all three datasets. This supports the general conclusion that ArbESC+ is better at handling errors that require a clear written signature than errors that require a deep understanding of context or semantics.

ArbESC+ is particularly robust to tags with stable surface references, but has difficulty dealing with learner errors that do not rely on stable writing patterns, highlighting the need for contextual and semantic enhancement of Arabic GEC models.

\begin{table*}[ht!]
\centering
\caption{ArbESC+ corrected percentages per ARETA tag on QALB-2014 and QALB-2015 (L1, L2).}
\small
\begin{tabular}{lccc}
\toprule
\textbf{Tag} & \textbf{QALB-2014 (\%)} & \textbf{QALB-2015 (L1) (\%)} & \textbf{QALB-2015 (L2) (\%)} \\
\midrule
PT & 96.8  & 95.2  & 91.7 \\
SF & 98.5  & 96.3  & 27.5 \\
MT & 62.5  & 50.0  & 35.2 \\
XT & 100.0 & 100.0 & 100.0 \\
OC & 94.0  & 52.8  & 33.6 \\
XF & 45.3  & 41.0  & 36.1 \\
OG & 92.4  & 88.5  & 79.6 \\
XG & 18.9  & 31.4  & 56.5 \\
MI & 51.7  & 42.7  & 43.2 \\
XN & 46.2  & 53.2  & 50.4 \\
OW & 85.9  & 85.7  & 55.6 \\
XM & 41.5  & 100.0 & 29.7 \\
SW & 100.0 & 69.9  & 47.3 \\
OA & 83.7  & 59.0  & 57.6 \\
OD & 66.2  & 68.4  & 55.2 \\
OM & 57.7  & 61.5  & 42.9 \\
SP & 100.0 & 100.0 & 100.0 \\
XC & 100.0 & 59.7  & 54.3 \\
OR & 54.9  & 61.1  & 59.6 \\
PC & 100.0 & 100.0 & 10.9 \\
OT & 95.9  & 69.6  & 84.4 \\
MG & 91.9  & 92.7  & 75.3 \\
OH & 91.7  & 73.1  & 38.0 \\
PM & 79.2  & 82.0  & 64.2 \\
\bottomrule
\end{tabular}
\label{tab:ArbESC_per_tag_percentages}
\end{table*}

%%%%%%%%%%%%%%%%%%%%%%%%%%%%%%%%%%%%%%%%%%%%%%%%%%%%%%%%%%%%%%%%%%%%%%%%%%%%%%%%%%%%%%%%%%%%%%%%%%%%%%%%%%%%%%%%%%%%%%%%%%%%%%%%%%%%%%%%%%%%%%%%%%%%%%%%%%%%%%%%%%%%%%%%%%%%%%%%%%%%%%%%%%%%%%%%%%%%%%%%%%%%%%%%%%%%%%%%%%%%%%%%%%%%%%%%%%%
A Table \ref{tab:overcount_detailed} shows the overcorrection rates (OCRs) for the PT and XT tags for the QALB14 and QALB15 (L1/L2) datasets. QALB14 and QALB15 (L1) show the highest OCR rates for the PT tag (unnecessary punctuation), and QALB15 (L2) shows the only slight decrease. As a result of learning a general statistical pattern that punctuation is related to sentence integrity, the model tends to include punctuation where the sentence is not needed, without sufficient understanding of the subtle context and stylistic boundaries surrounding punctuation and conjunctions in Arabic. This tag is therefore more sensitive to overcorrection than others.

The OCR rates for tag XT (deleting unnecessary words) are also high, but they tend to decline from QALB14 to QALB15 (L1) and then to QALB15 (L2). There is a greater correlation between unnecessary deletion errors in native speakers and those in learners based on stylistic differences. Due to the irregularity and lack of standard patterns found in learner texts (L2), deleting redundant words requires more contextual reasoning than relying solely on surface clues. The overcorrection rate in L2 is therefore lower than that in L1.

Overcorrection is not simply due to poor error recognition, but also to how the model represents stylistic norms, especially in tasks where surface linguistic clues alone are not sufficient for judgment. In order to reduce unnecessary editing decisions, especially for punctuation and deletion, OCR processing needs to incorporate a context-aware scorer or reranker model.

\begin{table*}[ht!]
\centering
\small
\caption{Over-Correction (OCR) details for frequent tags across QALB14 and QALB15 (L1/L2). OCR(\%) is computed as $100 \times \left(1 - \frac{\text{Fixed}}{\text{Raw}}\right)$.}
\label{tab:overcount_detailed}
\begin{tabular}{lccccccccc}
\toprule
\multirow{2}{*}{\textbf{Tag}} &
\multicolumn{3}{c}{\textbf{QALB14}} &
\multicolumn{3}{c}{\textbf{QALB15 (L1)}} &
\multicolumn{3}{c}{\textbf{QALB15 (L2)}} \\
\cmidrule(lr){2-4}\cmidrule(lr){5-7}\cmidrule(lr){8-10}
& \textbf{Raw} & \textbf{Fixed} & \textbf{OCR (\%)} 
& \textbf{Raw} & \textbf{Fixed} & \textbf{OCR (\%)} 
& \textbf{Raw} & \textbf{Fixed} & \textbf{OCR (\%)} \\
\midrule
PT (Unnec. punct.) & 746 & 1  & 99.87 & 560 & 1  & 99.82 & 543 & 13 & 97.60 \\
XT (Unnec. word)   & 252 & 17 & 93.25 & 166 & 18 & 89.16 & 253 & 46 & 81.82 \\
\bottomrule
\end{tabular}
\end{table*}

\section{Qualitative Comparison of Model Outputs}
\begin{figure*}[ht!]
\centering
\includegraphics[width=.85\textwidth]{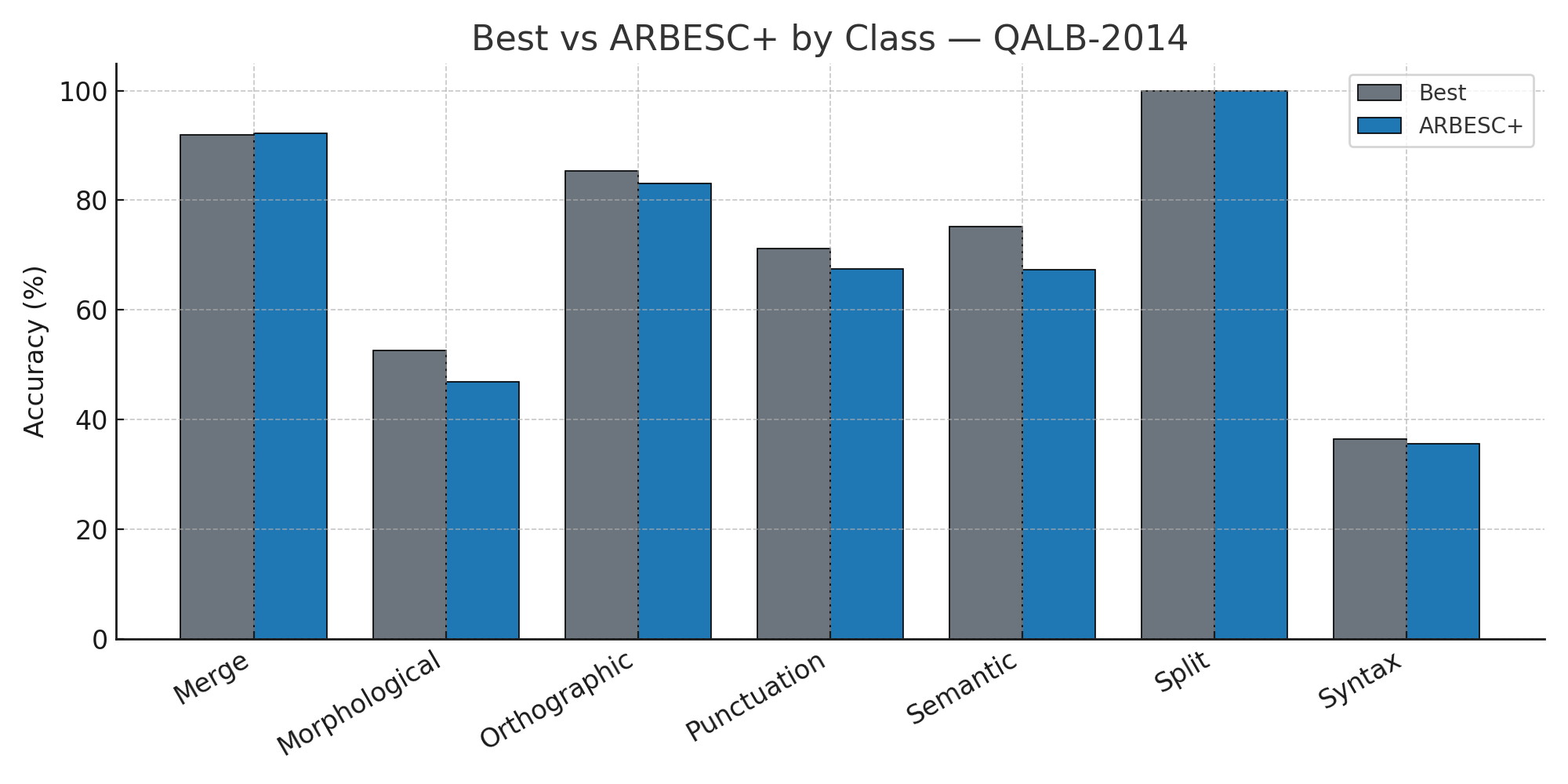}
\caption{Best vs. ARBESC+ by ARETA class — QALB-2014.}
\label{fig:best_vs_arbesc_q14}
\end{figure*}

\begin{figure*}[ht!]
\centering
\includegraphics[width=.85\textwidth]{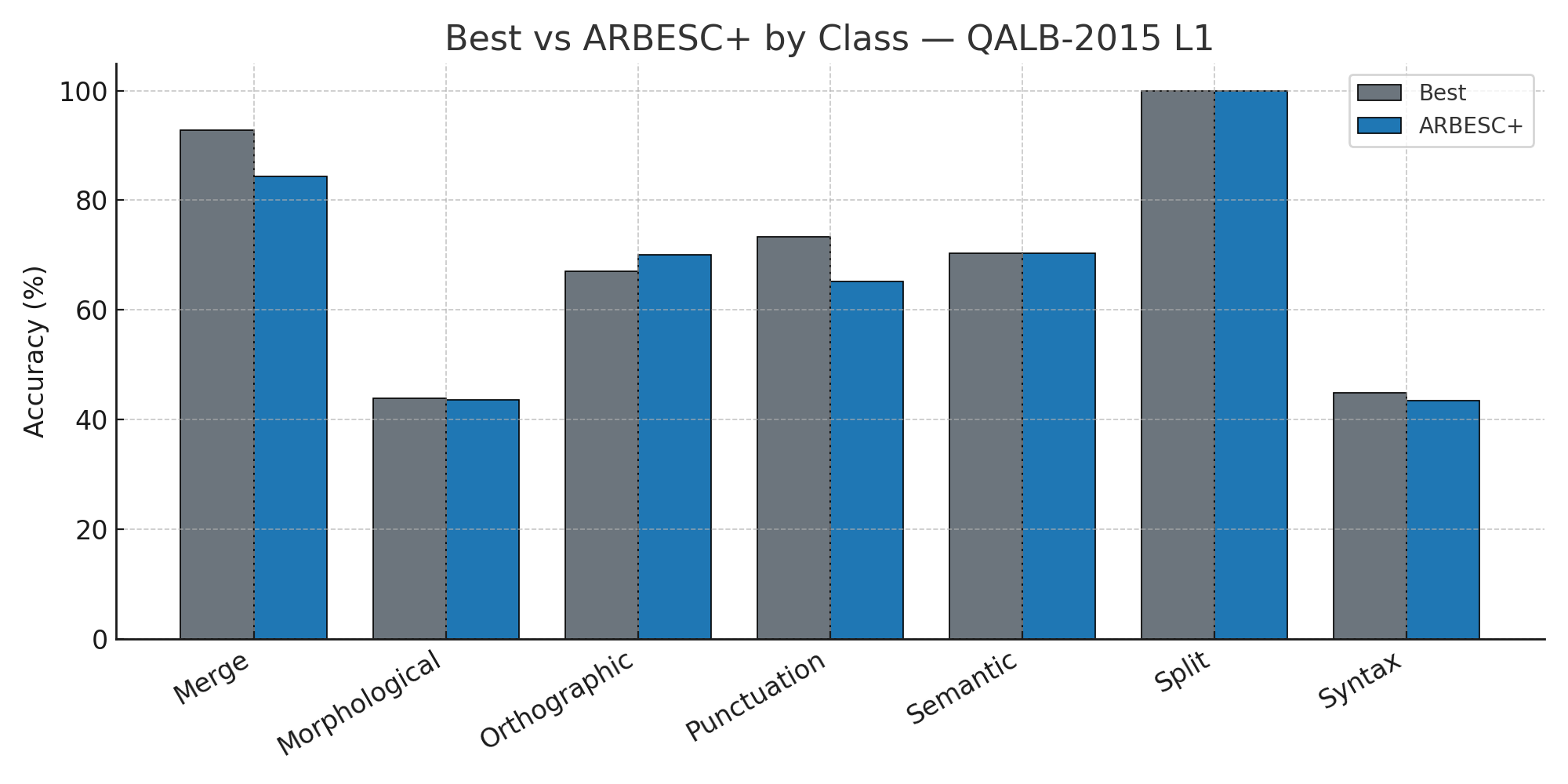}
\caption{Best vs. ARBESC+ by ARETA class — QALB-2015 L1.}
\label{fig:best_vs_arbesc_q15l1}
\end{figure*}

\begin{figure*}[ht!]
\centering
\includegraphics[width=.85\textwidth]{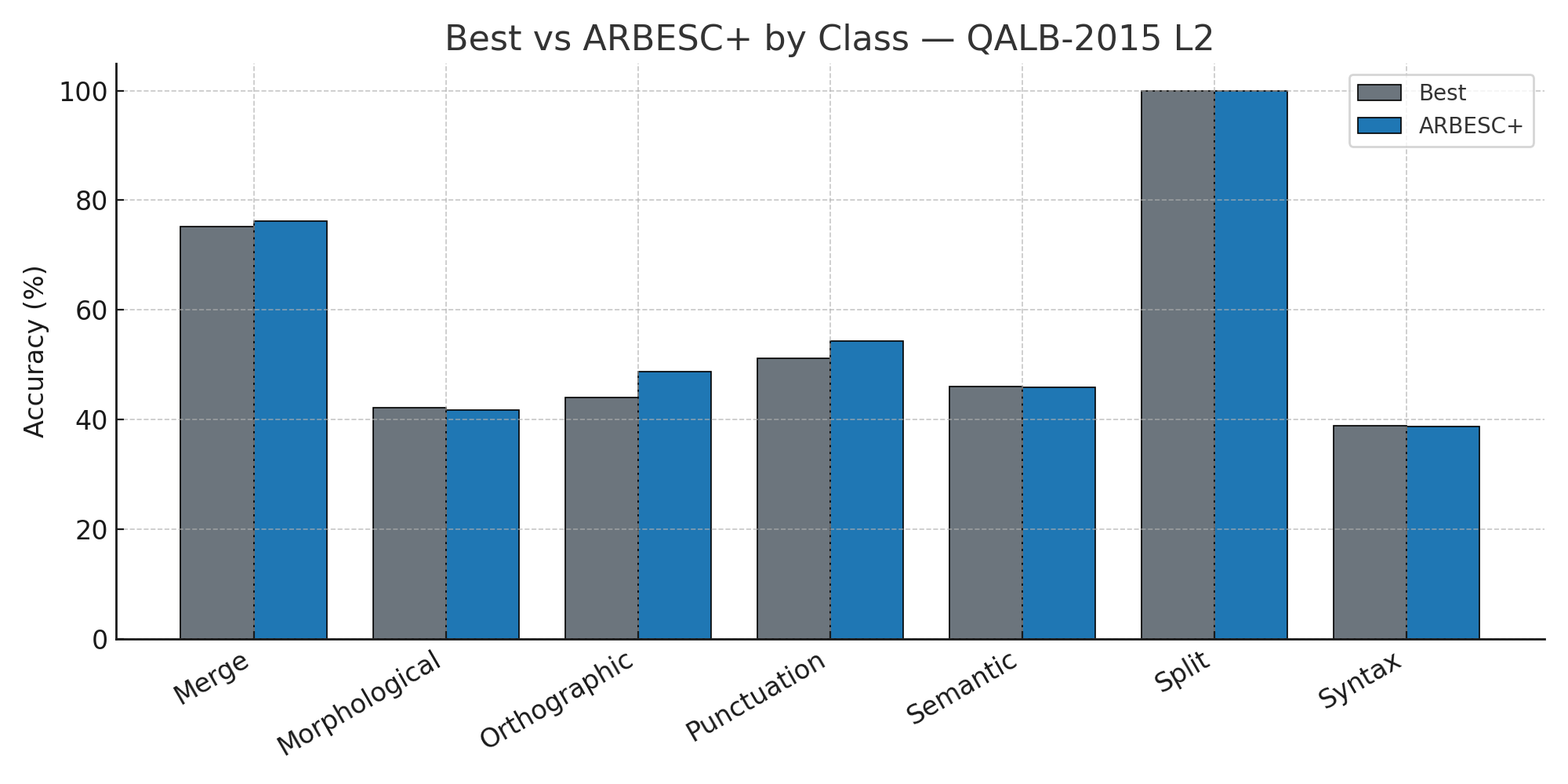}
\caption{Best vs. ARBESC+ by ARETA class — QALB-2015 L2.}
\label{fig:best_vs_arbesc_q15l2}
\end{figure*}

In the QALB-2014 and QALB-2015 test sets(L1 and L2), figures \ref{fig:best_vs_arbesc_q14}, \ref{fig:best_vs_arbesc_q15l1}, and \ref{fig:best_vs_arbesc_q15l2} show a direct comparison between the best model and ArbESC+ for the main linguistic error categories. There is a difference between each model's ability to handle different types of errors.

In all three figures, the two models maintain high and stable correction rates across all corpora in the Merge and Split categories. Due to their formal structure and consistent production rules, these errors are less context-dependent and more dependent on surface cues in the text to explain their stability.

As shown in Figure \ref{fig:best_vs_arbesc_q15l2}, the best model clearly outperforms ArbESC+ in Punctuation, Semantic, and Syntax categories, especially on QALB-2015 (L2) corpora. Consequently, the best model is better at handling contextual and semantic errors, whereas ArbESC+ relies more on localized correction strategies than syntactic relationships.

Additionally, Figure \ref{fig:best_vs_arbesc_q15l1} indicates that the gap between the two models is smaller in the L1 data than in the L2 data, indicating that the L2 errors are more diverse and random, requiring models based on deep linguistic understanding rather than simply retrieving training patterns. ArbESC+, however, performs competitively on orthographic and morphological errors when errors are regular and constrained within a learnable orthographic or derivational pattern shown in Figure \ref{fig:best_vs_arbesc_q14}.

ArbESC+ demonstrates that it is not simply a single-strategy correction system, but rather a framework that integrates multiple models and correction mechanisms. As a result, it was able to achieve a very close performance to the best model across many error categories, especially those characterized by clear writing patterns, such as mergers, segregation errors, and orthographic errors. The results confirm that Model Ensemble provides a greater ability to balance superficial and contextual modifications and reduces the system's sensitivity to text quality and source variations.

ArbESC+ does not differ significantly from the best model in error categories that require a deeper understanding of semantics or syntactics, although there is a slight gap between error categories that require a deeper understanding. Thus, ArbESC+ can be used for large-scale correction scenarios, especially when it is important to maintain reasonable training and operating costs while maintaining stable and general performance across various datasets.

%%%%%%%%%%%%%%%%%%%%%%%%%%%%%%%%%%%%%%%%%%%%%%%%%%%%%%%%%%%%%%%%%%%%%%%%%%%%%%%%%%%%%%%%%%%%%%%%%%%%%%%%%%%%%%%%%%%%%%%%%%%%%%%%%%%%%%%%%%%%%%%%%%%%%%%%%%%%%%%%%%%%%%%%%%%%%%%%%%%%%%%%5
\begin{figure}[ht] 
\centering 
\includegraphics[width=0.95\linewidth]{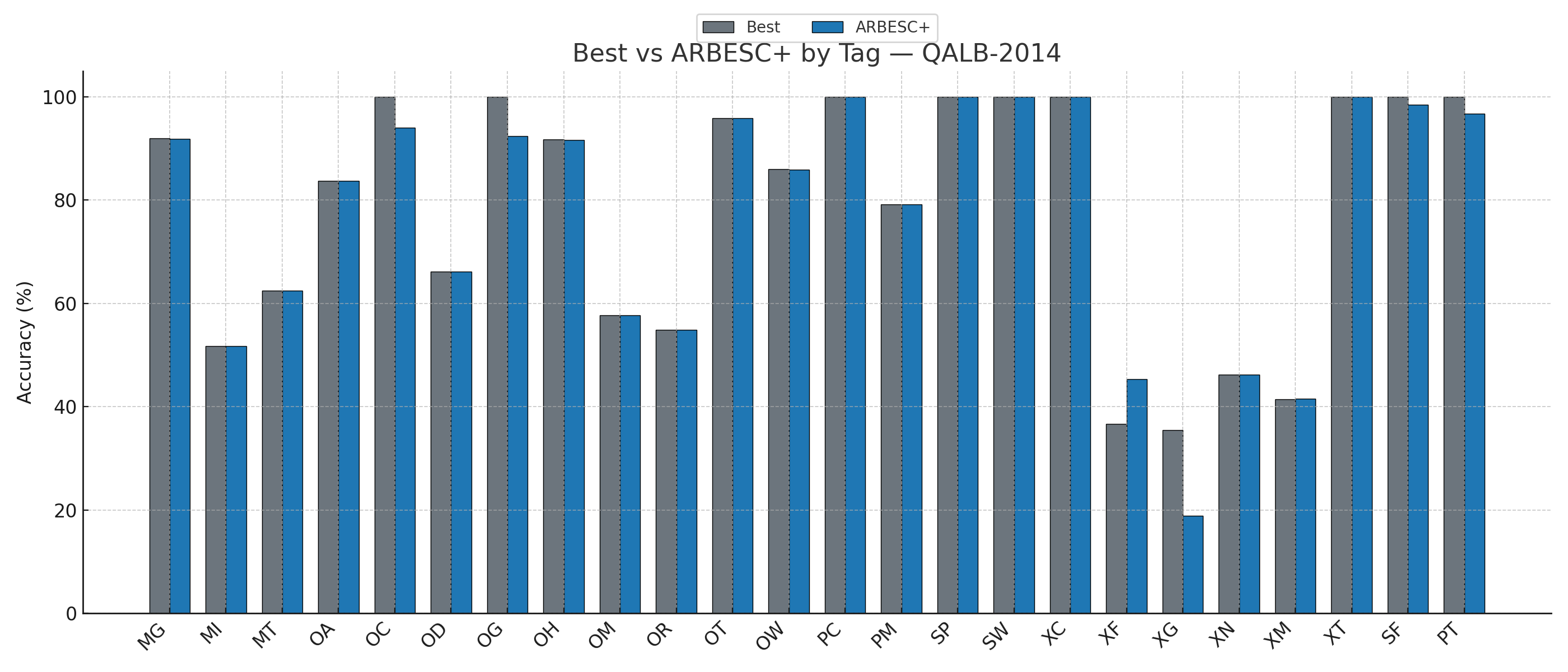} \caption{Best vs ARBESC+ performance by ARETA tag on QALB-2014. ARBESC+ closely matches the best model across most error tags, reflecting stable correction behavior on consistent native data.} \label{fig:qalb14_tags} \end{figure}

\begin{figure}[ht] 
\centering
\includegraphics[width=0.95\linewidth]{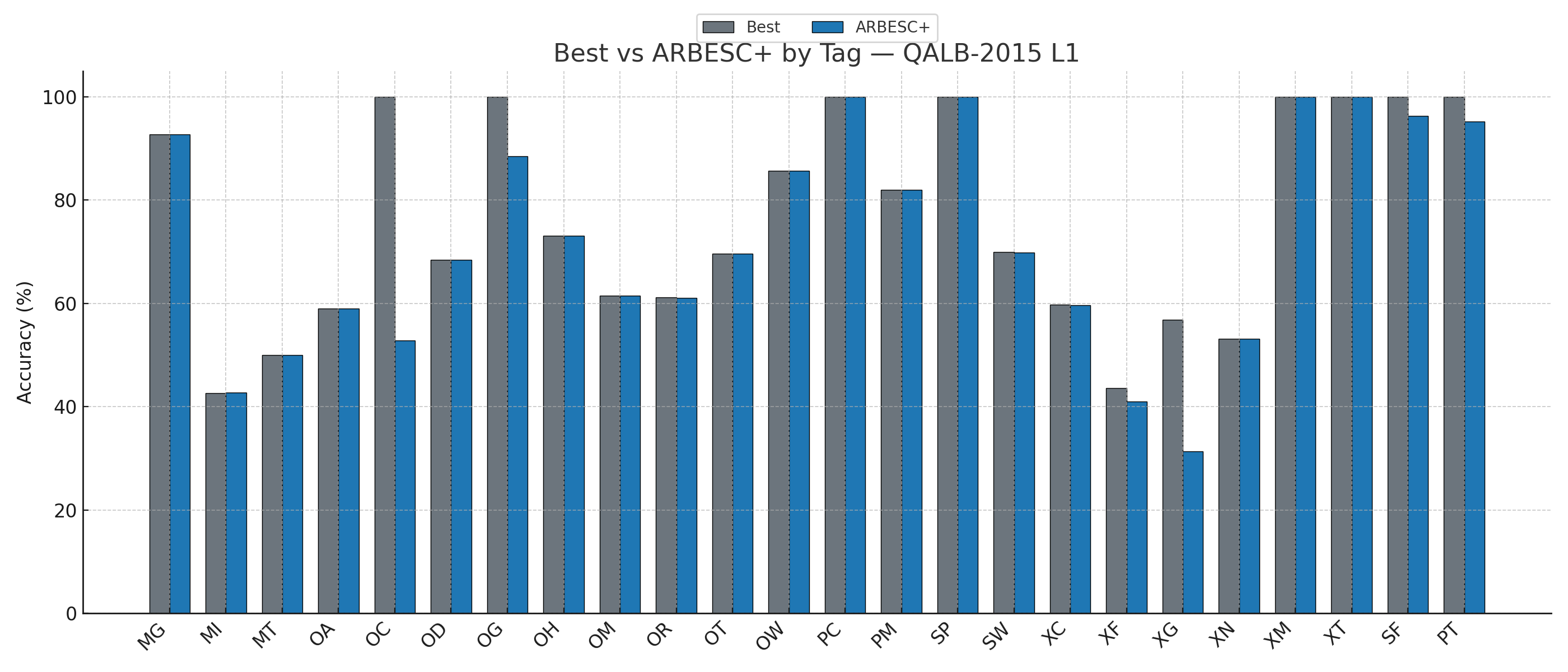} \caption{Best vs ARBESC+ performance by ARETA tag on QALB-2015 L1. The performance gap narrows on native speaker data, indicating more regular error patterns compared to L2.} \label{fig:qalb15_l1_tags} \end{figure}

\begin{figure}[ht] 
\centering \includegraphics[width=0.95\linewidth]{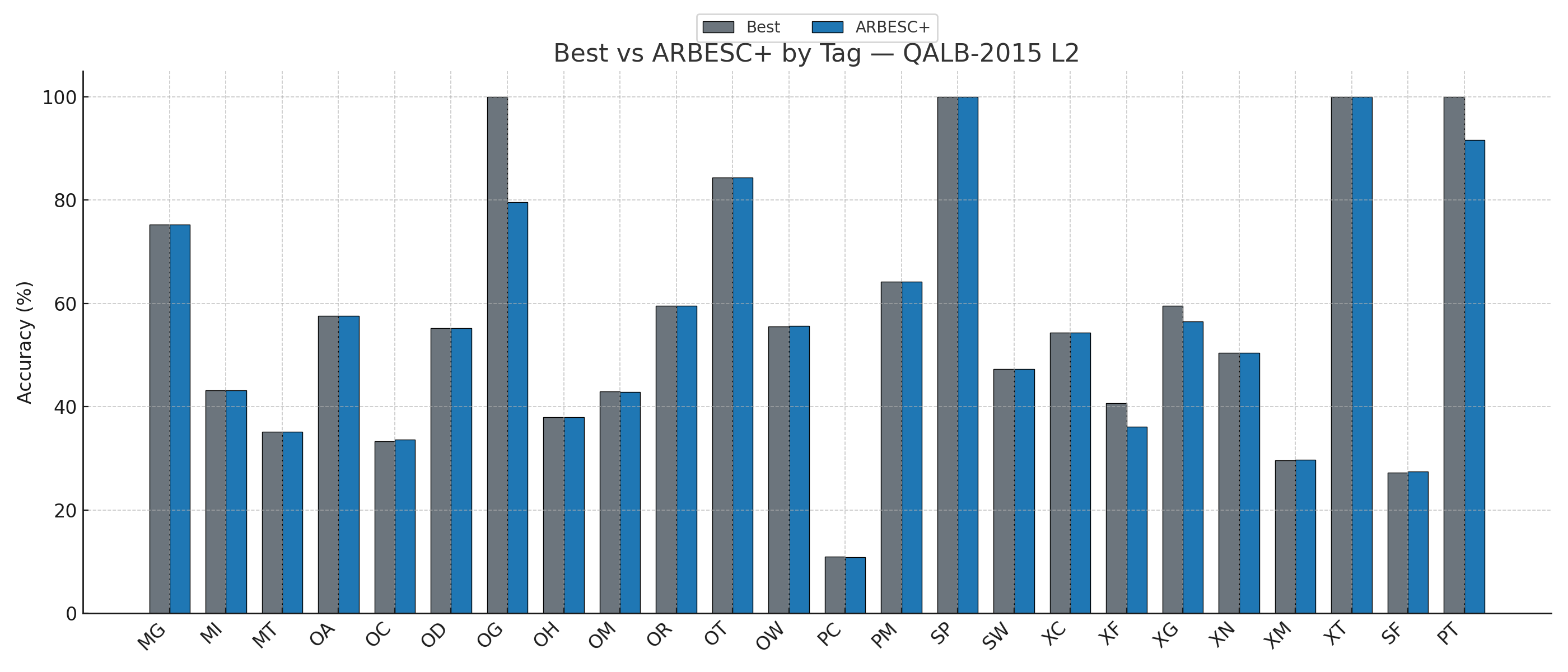}
\caption{Best vs ARBESC+ performance by ARETA tag on QALB-2015 L2. ARBESC+ approaches the best model on structural and orthographic tags but shows a wider gap in semantic and syntactic tags.} \label{fig:qalb15_l2_tags} 
\end{figure}

%%%%%%%%%%%%%%%%%%%%%%%%%%%%%%%%%%%%%%%%%%%%%%%%%%%%%%%%%%%%%%%%%%%%%%%%%%%%%%%%%%%%%%%%%%%%%%%%%%%%%%%%%%%%%%%%%%%%%%%%%%%%%%%%%%%%%%%%%%%%%%
As shown in \ref{fig:qalb14_tags}, \ref{fig:qalb15_l1_tags}, and \ref{fig:qalb15_l2_tags}, ArbESC+ performs better than the best model across the three QALB sets. A consistent pattern emerges across all constructs: ArbESC+ closely matches the best model's performance on tags requiring clear written cues or stable formal relationships, while the gap widens for tags requiring deeper grammatical understanding or semantic processing.

As shown in the Figure \ref{fig:qalb14_tags} for QALB-2014 data, ArbESC+ achieves a near-perfect match with the best model on most tags, especially those associated with spelling errors (OA, OG, OW) and merge and sever errors (MG, SP, XT). As native speaker data exhibits greater linguistic regularity and less variation in error forms, correction can be based on easily learned patterns.

The Figure \ref{fig:qalb15_l1_tags} for QALB-2015 L1 data shows a similar pattern with a slight variation. Despite slight differences at the tags related to context and meaning (XN, XG, OM), ArbESC+ still performs close to the best model at surface tags. There are still predictable errors among native speakers in this dataset, but they may differ in meaning depending on the context.

According to Figure \ref{fig:qalb15_l2_tags} for QALB-2015 L2), the gap widens for semantic and syntactic tags (e.g., XG, MT, OR), indicating higher randomness and lower structural regularity of L2 learner errors. 

Although ArbESC+ maintains good performance for both syntactic and orthographic tags, the Model Ensembling process allows it to maintain correction stability even when switching sources and linguistic backgrounds.

For most tags, ArbESC+ approaches the performance of the best model, especially those with clear formal patterns. The most prominent path for future improvement in L2 contexts remains to expand its semantic discrimination capabilities and process syntactic structure.


\begin{figure}[ht!]
\centering
\includegraphics[width=.95\linewidth]{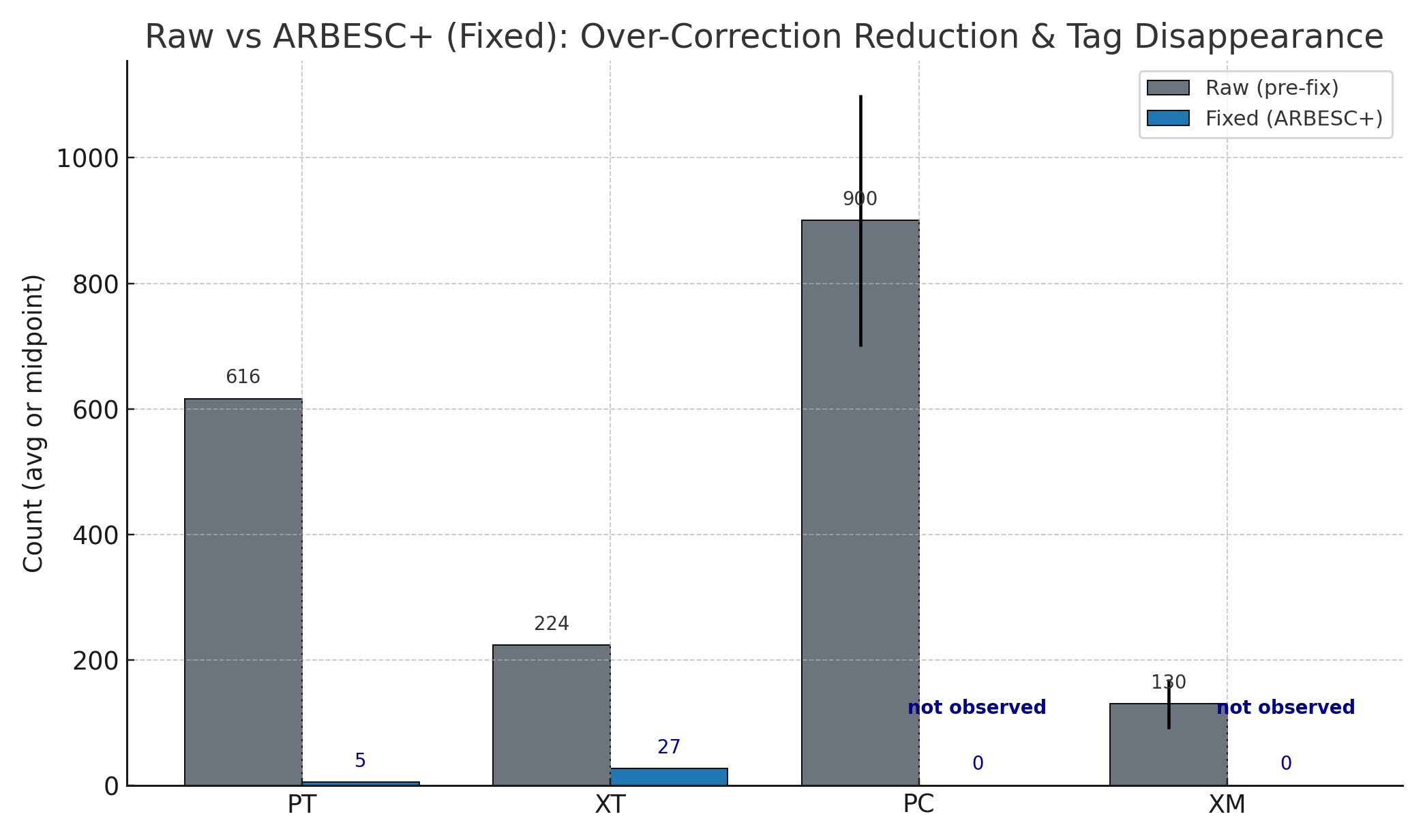}
\caption{Raw (gray) vs.\ ARBESC$+$ Fixed (blue) counts for frequent over-correction tags. PT/XT values are dataset averages; PC/XM use range midpoints (with error bars). PC and XM are not observed in the latest ARBESC$+$ run.}
\label{fig:arbesc_overcorr_single_all}
\end{figure}

%%%%%%%%%%%%%%%%%%%%%%%%%%%%%%%%%%%%%%%%%%%%%%%%%%%%%%%%%%%%%%%%%%%%%%%%%%%%%%%%%%%%%%%%%%%%%%%%%%%%%%%%%%%%%%%%%%%%%%%%%%%%%%%%%%%%%%%%%%%%%%%%%%%%%%%%%%%%%%%%%%%%%%%%%%%%%%%%%%%%%%%%%%%%%%
Finally, we compare the number of raw errors before correction to the number of errors left after correction for the following tags known for overcorrection: PT (unnecessary punctuation), XT (extra words), PC (confused punctuation), and XM (unjustified word additions or repetitions) as shown in figure. \ref{fig:arbesc_overcorr_single_all}.

With ARBESC+, the number of errors in all four tags is dramatically reduced, with PT and XT values dropping to very low levels, while PC and XM phenomena completely disappear. Hence, the model not only corrects basic errors, but also adjusts the correction thresholds so as to avoid overcorrection, a prominent problem in generative correction models.

Additionally, the model's stylistic sensitivity to punctuation and word functions within a sentence has improved, contributing to this decrease. ARBESC+ now distinguishes between what should be corrected and what should be retained. As a result, the model is more stable and usable in realistic contexts that require preserving the writer's style and original sentence structure.
The figure \ref{fig:arbesc_overcorr_single_all} shows that ARBESC+ not only achieved higher accuracy, but also gained the ability to control the level of interference, which significantly reduced overcorrection, a key factor in the quality of models applied to natural language.
\section{Conclusion}
The study presents the first integrated system for Arabic language error correction across all languages. With the framework, several models were combined, and the proposed corrections were classified according to statistical criteria supported by lightweight machine learning mechanisms, thereby improving accuracy and achieving a better balance between precision and recall. According to the results, the system integration clearly outperformed individual models, proving the efficacy of the proposed strategy and its importance for developing Arabic text correction tools.

\section*{Acknowledgements}
The Aziz Supercomputer at King Abdulaziz University was used for all experiments; therefore, we are grateful to the university's High-Performance Computing Center (HPC).

%Bibliography
\bibliographystyle{plainnat}
\bibliography{references}
\end{document}